\documentclass[10pt,twocolumn,letterpaper]{article}
\usepackage[pagenumbers]{cvpr} 
\usepackage{graphicx}
\usepackage{amsmath}
\usepackage{amssymb}
\usepackage{booktabs}
\usepackage{times}
\usepackage{epsfig}
\usepackage{subfigure}
\usepackage{tabu}
\usepackage{multirow}
\usepackage[numbers]{natbib}
\usepackage{stfloats}
\usepackage{colortbl}
\usepackage{xcolor}
\definecolor{citecolor}{RGB}{34,139,34} 
\usepackage[pagebackref,breaklinks,colorlinks,
citecolor=citecolor,bookmarks=false]{hyperref}
\usepackage[capitalize]{cleveref}

\crefname{section}{Sec.}{Secs.}
\Crefname{section}{Section}{Sections}
\Crefname{table}{Table}{Tables}
\crefname{table}{Tab.}{Tabs.}

\newlength\savewidth\newcommand\shline{\noalign{\global\savewidth\arrayrulewidth\global\arrayrulewidth 1pt}\hline\noalign{\global\arrayrulewidth\savewidth}}

\def\cvprPaperID{xxxx}
\def\confName{CVPR}
\def\confYear{2023}

\begin{document}

\title{An Empirical Study of Pseudo-Labeling for Image-based 3D Object Detection} 

\author{Xinzhu Ma$^1$ \ Yuan Meng$^2$ \ Yinmin Zhang$^1$ \ Lei Bai$^3$ \ Jun Hou$^4$ \ Shuai Yi$^4$ \ Wanli Ouyang$^1$ \\ 
$^1$ University of Sydney \quad $^2$ Tsinghua University \quad $^3$ Shanghai AI Lab \quad $^4$ SenseTime \\
\texttt{\small\{xinzhu.ma, yinmin.zhang, wanli.ouyang\}@sydney.edu.au} \\ 
\texttt{\small yuanmeng@mail.tsinghua.edu.cn} \quad \texttt{\small bailei@pjlab.org.cn} }

\maketitle

\begin{abstract} 
Image-based 3D detection is an indispensable component of the perception system for autonomous driving. However, it still suffers from the unsatisfying performance, one of the main reasons for which is the limited training data. Unfortunately, annotating the objects in the 3D space is extremely time/resource-consuming, which makes it hard to extend the training set arbitrarily. In this work, we focus on the semi-supervised manner and explore the feasibility of a cheaper alternative, i.e. pseudo-labeling, to leverage the unlabeled data. For this purpose, we conduct extensive experiments to investigate whether the pseudo-labels can provide effective supervision for the baseline models under varying settings. The experimental results not only demonstrate the effectiveness of the pseudo-labeling mechanism for image-based 3D detection (e.g. under monocular setting, we achieve 20.23 AP for moderate level on the KITTI-3D testing set without bells and whistles, improving the baseline model by 6.03 AP), but also show several interesting and noteworthy findings (e.g. the models trained with pseudo-labels perform better than that trained with ground-truth annotations based on the same training data). We hope this work can provide insights for the image-based 3D detection community under a semi-supervised setting. The codes, pseudo-labels, and pre-trained models will be publicly available.
\end{abstract}
\renewcommand{\thefootnote}{}
\footnote{\noindent work in progress.}

\section{Introduction}
As a crucial component of the self-driving system \cite{kitti,nuscenes,waymo}, 3D object detection has attracted extensive attention from both academia and industry. Especially, image-based 3D detection \cite{3dodi} has gradually become a hot problem in recent years. However, although lots of breakthroughs \cite{multifusion,pseudolidar,am3d,m3drpn,monodis,monodle,caddn,are_we,dd3d,gupnet,patchnet,stereorcnn,monopair,ligastereo,dsgn,monodistill,monoef,disprcnn_j,pseudolidar++} have been made, the performance of the image-based methods still significantly lags behind that of LiDAR-based methods, such as  \cite{pointpillars,pvrcnn,voxelrcnn,mv3d,voxelnet,pointrcnn}, and one of the main reasons for the unsatisfying performances of these methods is the limited training samples.

\begin{figure*}[t]
\centering
\includegraphics[width=1.0\linewidth]{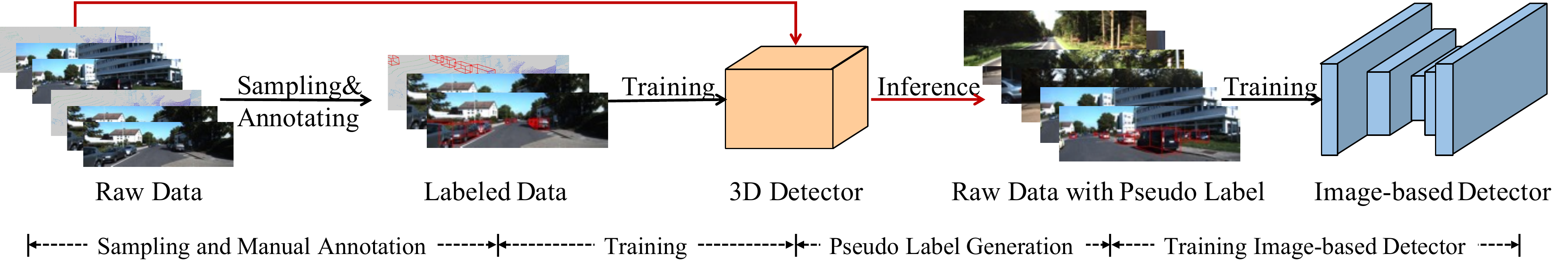}
\caption{{\bf The proposed pipeline for training an image-based 3D detector.} After collecting the raw data, the empirical practice is to sample some \emph{key-frames} and annotate them to generate the training data. Based on these training samples, we further train the 3D detectors and generate pseudo-labels for remaining unlabeled data. Finally, we use \emph{all} frames to train our models.}
\label{fig:framework}
\end{figure*}

Unfortunately, although the raw data is relatively easy to collect, manually annotating the objects in the 3D space is a complicated and labor-consuming task. To seek a cheap alternative to the manually annotate labels, we investigate whether the pseudo-labels can provide effective supervision for the image-based 3D object detectors. Particularly, as shown in Figure \ref{fig:framework}, we adopt the following paradigm to train the image-based 3D object detectors: (i) train a teacher model with the annotated key-frames; (ii) generating the pseudo-labels for the unlabeled data using the well-trained teacher model; (iii) training the image-based 3D detectors with the resulting pseudo-labels. Fortunately, our exploratory experiments reveal that the pseudo-labels can play the role of supervisor well, which encourages us to investigate it under more settings and going deeper for this mechanism. 

Specifically, we first adopt the LiDAR-based methods \cite{pvrcnn,voxelrcnn} as the pseudo-label generators. We argue this choice is meaningful because the LiDAR sweeps are required in the annotation process to provide 3D coordinates of the objects (and applying LiDAR points in the training phase is a common practise for existing image-based models, such as \cite{pseudo-label,caddn,monodistill,ligastereo,monorun,monopsr,pseudolidar++,dsgn,pseudolidare2e}). By this way, we demonstrate that the pseudo-labels generated from LiDAR-based models perform well in image-based 3D detection task, and existing models can be further improved by introducing more training samples. More interestingly, based on the same training samples, the models trained with the pseudo-labels significantly outperform these trained with the manual annotations. This counter-intuitive result suggests the promising application potential of pseudo-labeling in the field of image-based 3D detection, and we provide the empirical interpretation of this phenomenon.

Besides, we apply the pseudo-labeling approach on varying settings, {\it e.g.} semi-supervised learning with few annotated samples, and significantly surpasses current state of the art (SOTA) for most of them. Note that almost all the leading image-based models such as \cite{caddn,dsgn,monodistill,ligastereo} leverage LiDAR signals in their training phase, and we can build a fair environment to compare with them. The experimental results demonstrate our method still superior to these works in performance.  Furthermore, we also study whether the pseudo-label mechanism still works without LiDAR sweeps or better 3D detection models. Encouragingly, we find the models can also be effectively supervised by the knowledge they have learned before. In particular, take the monocular model GUPNet\cite{gupnet} as an example, we first train this model on the annotated frames, and then generate the pseudo-labels for the interframe sequences. After that, our model can benefited from the enlarged training set.

In summary, we provide an empirical study of the pseudo-labeling mechanism for image-based 3D object detection. In this work, we show that the pseudo-labeling can significantly improve the performance of existing image-based 3D detectors under varying settings. This simple approach can cooperatively work to almost all current SOTA methods and then provide new baselines for this community. Besides, we provide extensive experiments on the KITTI-3D benchmark \cite{kitti} to show the pseudo-labeling scheme in all-around, and the promising results firmly demonstrate the effectiveness of our method.

\section{Related Work}

\noindent
\textbf{Image-only-based 3D object detection.}
In the standard setting, only the RGB images, camera parameters, and object annotations are available for the image-based 3D detection task. However, even for the powerful convolutional neural networks (CNNs), it is extremely hard to build the mapping from 2D images to 3D bounding boxes based on such limited information. In order to better learn this mapping, the existing methods proposed for this setting mainly based on:  geometric priors \cite{m3drpn,monopair,deep3dbox,gs3d,monogrnet,rtm3d,tlnet,gupnet,monopair,monoflex,monogeo,monorcnn}, novel network designs \cite{oftnet,m3drpn,m3dssd,dsgn,stereorcnn,groomednms}, and better loss formulations \cite{monodis,monodle}. Different from these methods, our method choose to provide more samples with supervision to help the CNNs to find the hidden patterns. In this work, we use some representive models proposed for this setting as our baseline to show the effectiveness of our strategy. 

\noindent
\textbf{Image-based 3D detection with extra data.}
An effective method to improve the visual 3D detectors in performance is introducing extra data.
Specifically, \cite{multifusion,d4lcn,ddmp3d,dfrnet} choose use an depth estimator to predict the depth maps and then use them to augment the RGB images.
\cite{pseudolidar,am3d,monowithpl,decoupled3d,patchnet,demystifying,decoupled3d} adopt the `pseudo-LiDAR' paradigm, which transform the estimated depth maps into point clouds and then apply the LiDAR-based 3D detection methods on the resulting data.
Note that the there is no depth annotation in the provided data for this task, and the ground-truth depth maps are generally derived from the LiDAR sweeps by themselves or the KTITT team \cite{kitti}. 
Besides, some methods \cite{monorun,dsgn,ligastereo,caddn} also use LiDAR points to supervise their proposed networks/modules.
In summary, these works use LiDAR sweeps as extra annotations (or to generate geometry features) for \emph{existing samples}, while we use LiDAR sweeps to generate annotations for \emph{more samples}. Beside, we argue the proposed strategy is orthogonal to these works, and can work collaboratively with them.

\noindent
\textbf{Pseudo-labeling mechanism \cite{pseudo-label} for image-based 3D detection.}
As as a common semi-supervised learning method, pseudo-labeling has been successfully applied in various computer vision tasks, such as image classification, object detection, semantic segmentation. Recently, an arXiv paper also aware of the pseudo-labeling can be applied into monocular 3D detection. In particular, Peng {\it et al.} \cite{pl_mono} apply the this strategy on monocular 3D detection. However, this work only present a  preliminary attempt for this problem at one setting. By contrast, we make a deeper exploration, provide in-depth analysis, discuss this problem in novel perspective and varting settings.

\noindent
{\bf Semi-supervised learning for image-based 3D Detection.}
As a promising research direction, semi-supervised learning (SSL) attracted lots of attentions. For 3D detection, the annotation resource is valuable and there are lots of unlabeled data, which makes this task very suitable for applying SSL methods.
Unfortunately, although several works \cite{ioumatch,sess,sessd} have introduce SSL to LiDAR-based 3D detection, little work \cite{mono_ssl} has been done to discuss this problem for image-based models. In this work, we leverage the LiDAR-based methods' enormous potential to generate pseudo-labels for the unlabeled frames, and train the monocular/stereo 3D detectors on the unlabeled dataset without ground truth while obtaining better performance over the state-of-the-art monocular/stereo detectors trained on the labeled dataset.
\section{Approach}

The objective of this work is to study the pseudo-label mechanism \cite{pseudo-label} in image-based 3D detection. As shown in Figure \ref{fig:framework}, the whole pipeline can be clearly divided into two parts: generating pseudo-labels using teacher models and training the baseline models with the resulting pseudo-labels. Next, we detail the settings we used in this work, including the pseudo-label generators, baseline models, datasets, etc.

\noindent
{\bf Datasets and metrics.}
We conduct the experiments on the most commonly used KITTI-3D dataset \cite{kitti}, which provides 7,481 annotated frames for training and 7,518 frames for testing. Following \cite{mono3d,mv3d}, we split the 7,481 training samples into a training set (3,712 frames) and a validation set (3,769 frames). Besides, we also use the KITTI raw data, which provides 155 video sequences, with about 48K unlabelled frames in total. Note that these frames include the $\sim$7K frames in KITTI-3D's training set, while the raw data for the testing set is not released.  We further split the raw data into several sub-sets to evaluate the pseudo-labeling approach under varying settings. The summary of these splits is presented in Table \ref{table:splits}. Besides, some sub-sets of these splits are used to show the performance changes \emph{w.r.t.} the size of training data, which will be further explained in the corresponding experiment parts.
As for the metrics, following \cite{monodis}, we evaluate the models with the $\rm{AP}|_{R_{40}}$ for both 3D detection and Bird's Eye View (BEV) detection tasks.
We mainly focus on the Car category, and both 0.7 and 0.5 IoU thresholds are considered.
The performance of Pedestrian and Cyclist is also reported for reference.

\begin{table}[h]
\centering
\resizebox{\linewidth}{!}{
\setlength\tabcolsep{4.00pt}
\begin{tabular}{l|ccccc}
 ~ & train & val & eigen & eigen-clean & all \\ 
\shline
\# annotated & 3,712 & 3,769 & - & - & 7,481 \\
\# total & 13,596 & 10,670 & 23,488 & 14,940 & 47,937 \\
\end{tabular}}
\caption{\small{{\bf Summary of the data splits.} We generate train/val split by collecting all the frames from the video sequences which correspond to the KITTI 3D's training/validation images. Eigen denotes Eigen's training set \cite{eigen} which is commonly used in the KITTI Depth benchmark. Following \cite{are_we}, we generate the eigen-clean split by removing the images geometrically close to the images in the KITTI 3D's validation set.}}
\label{table:splits}
\end{table}

\noindent
{\bf Baseline models.}
To ensure the generality and reproducibility, we choose some recently published methods with official codes as our baselines.
In particular, we choose two monocular 3D detectors (\emph{i.e.}, one-stage MonoDLE \cite{monodle} and two-stage GUPNet \cite{gupnet}) which only need images and camera parameters in both training and inference phases as our baseline models.
For the stereo setting, we use the LIGA-Stereo \cite{ligastereo} in our experiments. Note that LiDAR points are required in the training phase for this baseline. Thus, we also build a pure stereo baseline by removing the relevant requirements (\emph{i.e.}, depth loss on the cost-volume and the cross-modality knowledge distillation losses), marked as \underline{LIGA-Stereo} in our experiments. More details about the baseline models can be found in Appendix \ref{appendix:baselines}.

\noindent
{\bf Pseudo-label generators.}
We first adopt two LiDAR-based models (PV R-CNN \cite{pvrcnn} and Voxel R-CNN \cite{voxelrcnn}) to generate the pseudo-labels. Although this makes the LiDAR points involve in the training phase, we argue this strategy is meaningful, mainly based on the following two considerations:
\begin{itemize}
	\item annotating objects in the 3D space requires the accurate spatial information, which is usually provided by LiDAR points. Therefore, the training of the teacher models is hard to completely avoid the involve of LiDAR points.
	\item almost all SOTA image-based 3D detectors adopt LiDAR points as supervision in the training phase, so we also test our method under this setting for a fair comparison.
\end{itemize}
Besides, to explore the monocular/stereo only setting, we also use GUPNet and \underline{LIGA-Stereo} as the pseudo-label generators. Note that using original LiDAR-Stereo to generate pseudo-labels is lack of practical significance, because training LIGA-Stereo also requires LiDAR points, and we can directly use LiDAR detectors to generate pseudo-labels at this setting. 

\noindent
{\bf Implementation.}
We adopt the same hyper-parameters and training protocols in \cite{monodle,gupnet,ligastereo,voxelrcnn,pvrcnn} unless otherwise stated. The codes, pseudo-labels, and pre-trained models will be released for the reproducibility.

\section{Experiments and Analysis}

\subsection{Quality of the Generated Pseudo-Labels}
\label{sec:quality}
The first problem is whether the pseudo-labels can be used as the training labels of image-based 3D detectors or not. If yes, what is the quality of the pseudo-labels compared with the official annotations? To investigate these problems, we first generate the pseudo-labels for the official training images and then compare the performances of the image-based 3D detectors trained from these pseudo-labels or official annotations.

\begin{table*}[!t]
\centering
\resizebox{0.8\linewidth}{!}{
\setlength\tabcolsep{5.00pt}
\begin{tabular}{r|ccc|ccc|ccc|ccc}
\multirow{2}{*}{~} & \multicolumn{3}{c|}{MonoDLE} & \multicolumn{3}{c|}{GUPNet} & \multicolumn{3}{c|}{\underline{LIGA-Stereo}} & \multicolumn{3}{c}{LiGA Stereo} \\
~ &  Easy & Mod. & Hard & Easy & Mod. & Hard & Easy & Mod. & Hard & Easy & Mod. & Hard \\ 
\shline
PV R-CNN & 19.32  & 14.96 & 13.45  & 23.24  & 17.37  & 15.58  & -  & -  & -  & -  & -  & - \\
Voxel R-CNN  & 20.34 & 15.78 & 13.82 & 24.99 & 18.10 & 16.22 & 77.21 & 58.67 & 55.84 & 83.85 & 66.40 & 63.23 \\
GT & 17.97 & 14.30 & 12.18 & 21.88 & 15.80 & 13.23 & 75.82 & 57.53 & 54.09
& 81.18 & 64.58 & 59.45\\
\hline
PV R-CNN & 60.32 & 44.98 & 41.02 & 63.00 & 47.65 & 42.42 & -  & -  & - & -  & -  & - \\
Voxel R-CNN  & 59.71 & 46.38 & 42.69 & 64.98 & 48.83 & 44.89 & 96.64	& 88.43	& 80.28 & 97.02	& 89.87	& 87.94 \\
GT & 57.88 & 44.03 & 39.40 & 58.99 & 43.85 & 38.94 & 94.80  & 87.58	& 79.92
& 96.77	& 89.59	& 87.60\\
\end{tabular}}
\caption{\small{\textbf{Quality of the pseudo-labels.} We report the performances of four baseline models trained from the pseudo-labels generated by two LiDAR-based detectors (PV R-CNN and Voxel R-CNN).
Metrics are the $\rm{AP}|_{R_{40}}$ for 3D detection with 0.7 (\emph{upper group}) and 0.5 IoU thresholds (\emph{lower group}).
We also show the performances of the models trained from the ground truth (GT) for reference.
All the baselines are trained on the 3,712 \emph{training} images and evaluated on \emph{validation} set.}}
\label{tab:quality_pl}
\end{table*}
\noindent
{\bf First attempt on pseudo-labeling.} We first investigate whether the LiDAR-based 3D detectors, representing the best-performing 3D detectors, can generate good enough pseudo-labels. To avoid biased conclusions caused by the over-fitting of the pseudo-label generators on the training split, we train the LiDAR-based models on the KITTI \emph{validation} set, and generate the pseudo-labels for the \emph{training} set. After that, we train our baseline models with the generated labels. The summary of the experimental results are shown in Table \ref{tab:quality_pl}. From these results, we can observe that the pseudo-labels perform well in providing supervision, even better than the manually annotated labels (especially for the monocular baselines). This shows the promising potential of pseudo-labeling for image-based 3D detection.

Besides, note that the KITTI dataset captures the 3D points with a 64-beam LiDAR, and sparser LiDAR singals are also well applied in other datasets (\emph{e.g.} nuScenes \cite{nuscenes} and Argoverse \cite{argoverse} adopt the 32-beam LiDAR), thus we generate the simulated 32-beam and 16-beam LiDAR sweeps for further investigation, and the experiments for this part can be found in Appendix \ref{appendix:sparse_lidar}.

\begin{figure*}[t]
\begin{minipage}[c]{0.55\linewidth}
\centering
\includegraphics[width=0.8\linewidth]{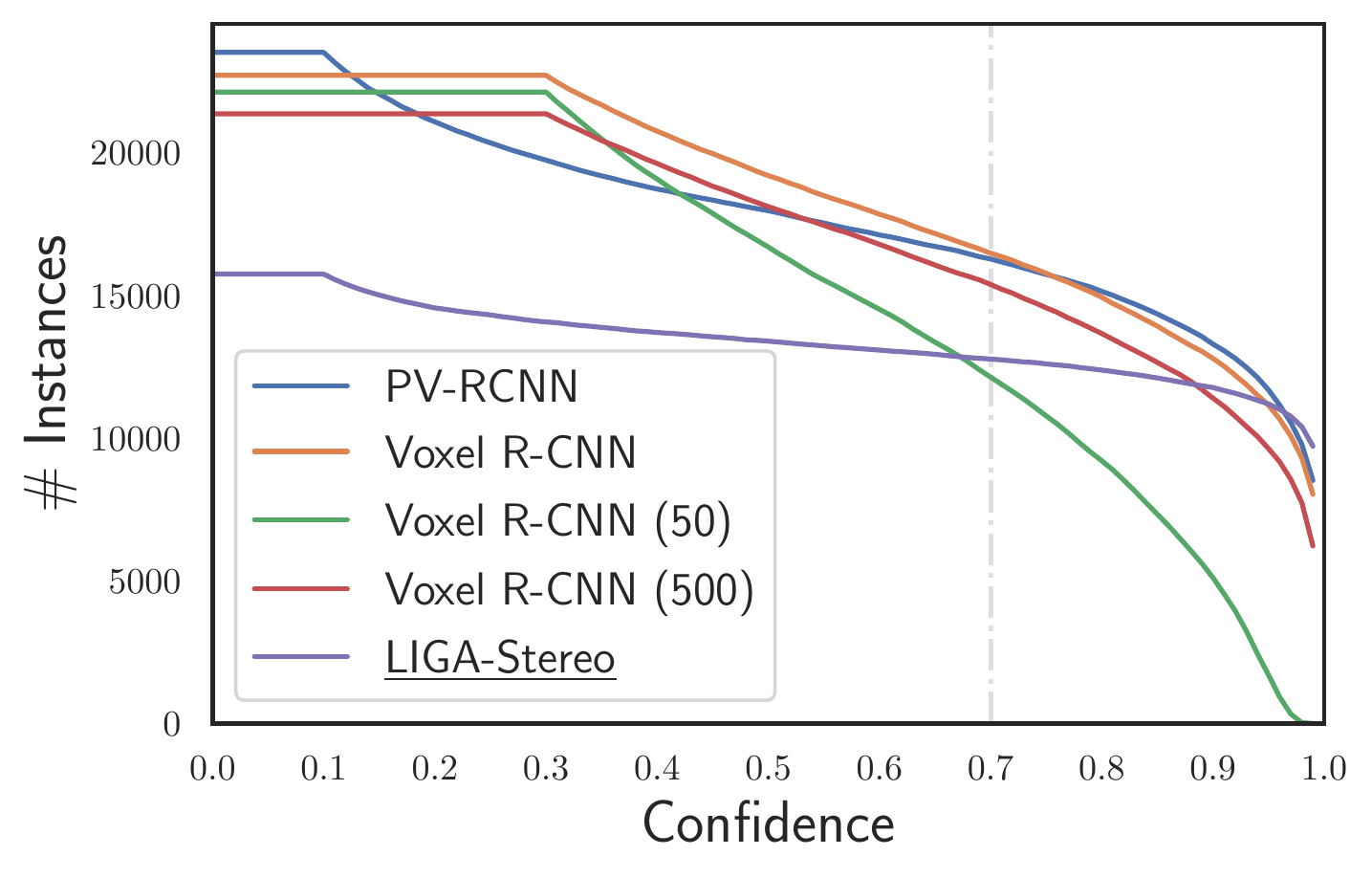}
\end{minipage}
\begin{minipage}[c]{0.45\linewidth}
\setlength\tabcolsep{5.00pt}
\resizebox{0.8\linewidth}{!}{
\centering
\begin{tabular}{lc|ccc}
method & threshold & Easy & Mod. & Hard  \\ 
\shline
MonoDLE & 0.5 & 21.42 & 16.26 & 14.87 \\
MonoDLE & 0.6 & 21.84 & 17.32 & 15.33 \\
MonoDLE & 0.7 & 22.14 & 16.97 & 15.56 \\
MonoDLE & 0.8 & 21.23 & 16.86 & 15.38 \\
\hline
GUPNet & 0.5 & 25.62 & 18.76 & 16.83 \\
GUPNet & 0.6 & 24.91 & 18.28 & 16.45 \\
GUPNet & 0.7 & 26.97 & 19.05 & 17.01 \\
GUPNet & 0.8 & 26.17 & 18.71 & 16.91 \\
\end{tabular}} 
\end{minipage}
\caption{\emph{Left:} the accumulated numbers (cars) of pseudo-labels with the confidence. \emph{Right:} threshold grid search for monocular detectors. We show the $\rm{AP}|_{R_{40}}$ under moderate setting with 0.7 IoU threshold.  The pseudo-label generators are trained on \emph{training} set, and the data are collected on \emph{validation}.}
\label{fig:thresholds}
\end{figure*}

\noindent
{\bf Removing low-quality pseudo-labels by confidence.} In the above experiments, we directly use the pseudo-labels to train our models. By analysing the detection results, we find there are some noisy samples in the results of the LiDAR-based 3D detectors. To remove the potential negative impact caused by these noises, we further filter the pseudo-labels by their confidences. Specifically, we conduct a small grid search on the confidence thresholds, and show the results in Figure  \ref{fig:thresholds}. We can find 0.7 threshold gives a good results, although it is not always the best, and we use this threshold for the following experiments in default. Meanwhile, we also show the accumulated instances along with confidences, which suggests about 18K car instances are kept at 0.7 confidence. Note that the annotations provide about 14K ground-truth cars for the same frames.

\begin{table}[t]
\centering
\setlength\tabcolsep{5.00pt}
\resizebox{\linewidth}{!}{
\begin{tabular}{l|lll|lll}
\multirow{2}{*}{~} &
\multicolumn{3}{c|}{$\rm{AP}|_{R_{40}}$@3D} & \multicolumn{3}{c}{$\rm{AP}|_{R_{40}}$@BEV} \\
method &  Easy & Mod. & Hard &  Easy & Mod. & Hard\\ 
\shline
VisualDet3D\cite{groundaware} &  23.63 & 16.16 & 12.06 & - & - & - \\
DLE\cite{dle} & 26.43 & 16.72 & 13.02 & 34.06 & 22.59 & 16.96 \\
SGM3D\cite{sgm3d} & 25.96 & 17.81 & 15.11 & 34.10 & 23.62 & 20.49 \\
\hline
Baseline (mono) & 21.88 & 15.80 & 13.23 & - & - & - \\
Ours & 27.72 & 19.38 & 17.11 & - & - & -
\end{tabular}}
\caption{\small{Comparison on the methods apply stereo images in the training phase. The data of upper group are from their papers. All models are trained from 3,712 frames.}}
\label{tab:stereo-guided}
\end{table}
\noindent
{\bf Can pseudo-labeling work without LiDAR points?}
We mainly discuss the LiDAR-based pseudo-labeling above, and the further problem is can pseudo-labeling work without LiDAR points? For this problem, we consider the following settings: (i) both the stereo and monocular images are available, and (ii) only the monocular images are available. For the first setting, we can use the stereo 3D detectors to generate the pseudo-labels for monocular models. This setting is also adopted by several methods \cite{dle,groundaware,rtm3d,sgm3d}, here we compare the proposed method to these works in Table \ref{tab:stereo-guided}. For a fair comparison, we do not introduce any other frames, and use the same data with pseudo-labels generated from \underline{LIGA-Stereo} to train our model. The experimental results show the stereo 3D detectors can also generate good pseudo-labels for monocular ones, and our method shows better performance than the competitors with the same setting (\cite{groundaware,dle} use stereo images to augment the training set while \cite{sgm3d} use stereo 3D detector to provide guidance for their model using knowledge distillation \cite{kd}).

As for the more challenging setting (ii), we find it can also work in some specific scenarios, and we will further discuss it in Section \ref{sec:discussion}.

\subsection{Scalability of the Training Samples}
\label{sec:scalability}
After confirming the effectiveness of the pseudo-labels, this section will provide  more applications and in-depth analysis of pseudo-labeling for image-based 3D detection.

\begin{table*}[t]
\centering
\setlength\tabcolsep{5.00pt}
\resizebox{0.8\linewidth}{!}{
\begin{tabular}{lc|lll|lll}
\multirow{2}{*}{~} &
\multirow{2}{*}{~} &
\multicolumn{3}{c|}{$\rm{AP}|_{R_{40}}$@IoU=0.7} & \multicolumn{3}{c}{$\rm{AP}|_{R_{40}}$@IoU=0.5} \\
split & \# images &  Easy & Mod. & Hard &  Easy & Mod. & Hard\\ 
\shline
a. training (gt) & 3,712 & 22.37$_{\pm0.65}$ & 16.21$_{\pm0.42}$ & 13.98$_{\pm0.63}$ & 60.50$_{\pm1.41}$ & 44.97$_{\pm1.25}$ & 39.99$_{\pm1.01}$ \\
b. training (pl) & 3,712 & 25.87$_{\pm1.39}$ & 19.07$_{\pm0.91}$ & 16.52$_{\pm0.88}$ & 64.30$_{\pm2.37}$ & 48.83$_{\pm1.63}$ & 44.31$_{\pm1.31}$ \\
c. train & 13,596 & 26.68$_{\pm0.34}$ & 19.92$_{\pm0.52}$ & 17.50$_{\pm0.20}$ & 66.36$_{\pm0.74}$ & 50.90$_{\pm0.64}$ & 46.58$_{\pm0.65}$ \\
d. eigen-clean  & 14,940 & 31.82$_{\pm0.71}$ & 23.49$_{\pm0.85}$ & 20.97$_{\pm0.49}$ & 67.65$_{\pm0.72}$ & 51.58$_{\pm0.58}$ & 48.04$_{\pm0.20}$ \\
e. train $\cup$ eigen-clean  & 28,536 & 35.77$_{\pm0.63}$ & 26.00$_{\pm0.54}$ & 22.69$_{\pm0.16}$ & 73.56$_{\pm0.69}$ & 57.55$_{\pm0.33}$ & 53.47$_{\pm0.27}$ \\
\hline
f. eigen  & 23,488 & 50.94 & 40.05 & 35.32 & 80.35 & 68.10 & 62.25 \\
g. eigen-sampling  & 14,940 & 50.85 & 38.56 & 34.75 & 79.92 & 65.94 & 60.12 \\ 
\end{tabular}}
\vspace{2mm}
\caption{\small{\textbf{More training samples.} Performances on KITTI \emph{validation} set of GUPNet trained from different data splits. $\pm$ captures the standard deviation over 5 runs. Pseudo-labels are generated by Voxel R-CNN trained from KITTI-3D's \emph{training} split.}}
\label{tab:more_samples}
\end{table*}

\noindent
{\bf Training with more samples.}
We first investigate whether the 3D detectors can benefit from more training samples. Particularly, we train the pseudo-label generators on the \emph{training} set and generate the pseudo-labels for the KITTI raw data. Then we divide the raw data into several splits (see Table \ref{table:splits} for more details of these data splits) and compare the performances of the selected baseline models trained from them. The experiments shown in Table \ref{tab:more_samples} indicates that the monocular 3D detectors can benefit from more training samples. For instances, 10K/25K more training data (b$\rightarrow$d and b$\rightarrow$e) can bring another 4.42/6.93 AP improvements (IoU=0.7, moderate setting).

\noindent
{\bf Stability.}
A common issue for existing monocular 3D detectors is the unstable performance, especially on the KITTI 3D benchmark. In this work, we find this problem can be largely alleviated by introducing more training samples, especially the temporal sequences. In particular, in the experiments presented by Table \ref{tab:more_samples}, we conduct multiple runs and report the mean and standard deviation for the unbiased comparison. From these results, we can find that (i): increasing the training samples (b$\rightarrow$d, d$\rightarrow$e) is helpful for improving the stability, along with the accuracy, of the algorithms; (ii): compared with simply extending the size of training set, introducing the temporal sequences (video cues) can better  improve the stability (b$\rightarrow$c). In summary, these two observations imply that introducing images with novel scenes tend to boost the models in performance, while more annotated samples in the same scenes can make the models more stable. 

\noindent
{\bf Data leakage.} Lots of monocular 3D detection methods adopted depth estimators, which are generally pre-trained on the KITTI raw data with Eigen's split \cite{eigen}, to provide depth cues for their models. However, there is an overlap between Eigen's training set and KITTI-3D's validation set. In particular, these two data splits share eight video sequences with 2,859 frames in total. Although the KITTI 3D validation set only include some key-frames among them, there are still 1,258 validation images overlap with the Eigen's training set. Here we quantitatively analyze the bias in performance caused by the data leakage. In particular, following \cite{are_we}, we remove the images, which are geographically close to any images in the validation set, and compare the performances of the models trained from this split (eigen-clean) and Eigen's training set (eigen). Furthermore, we also randomly sample the images from eigen split to generate a new training set (eigen-sampling), which has the same size as eigen-clean split, for a fairer comparison. As shown in Table \ref{tab:more_samples}, both the models trained on eigen and eigen-sampling splits show anomalous high performances, compared with that trained from eigen-clean, which suggests the data leakage will cause seriously unfair performance comparison on \emph{validation} set.

\begin{figure*}[t]
\centering\includegraphics[width=1.0\linewidth]{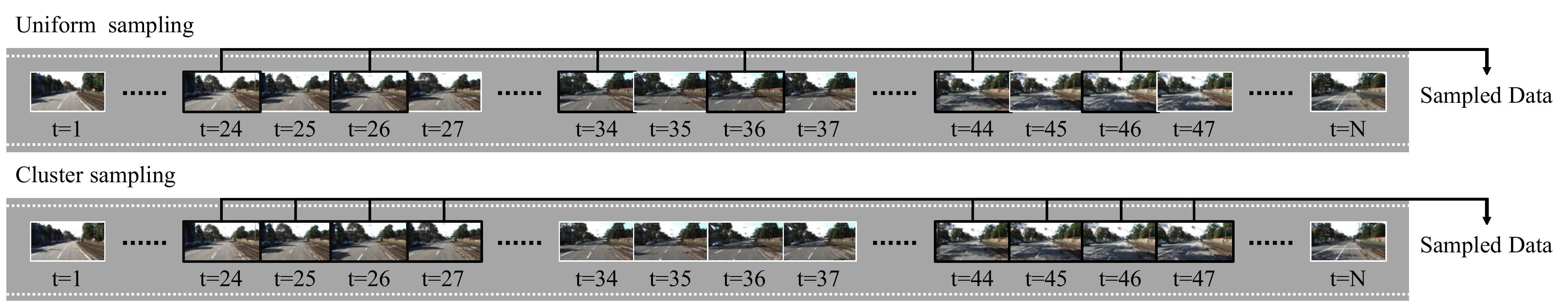}
\caption{Illustration of the sampling strategies.
\emph{Top}: Uniform sampling, we use this sampling scheme to simulate the different frame rates during data collection.
\emph{Bottom}: Cluster sampling, we use this sampling scheme to divide images into several clips.}
\label{fig:sampling}
\end{figure*}

\begin{figure*}[t]
\centering
{\includegraphics[width=0.4\linewidth]{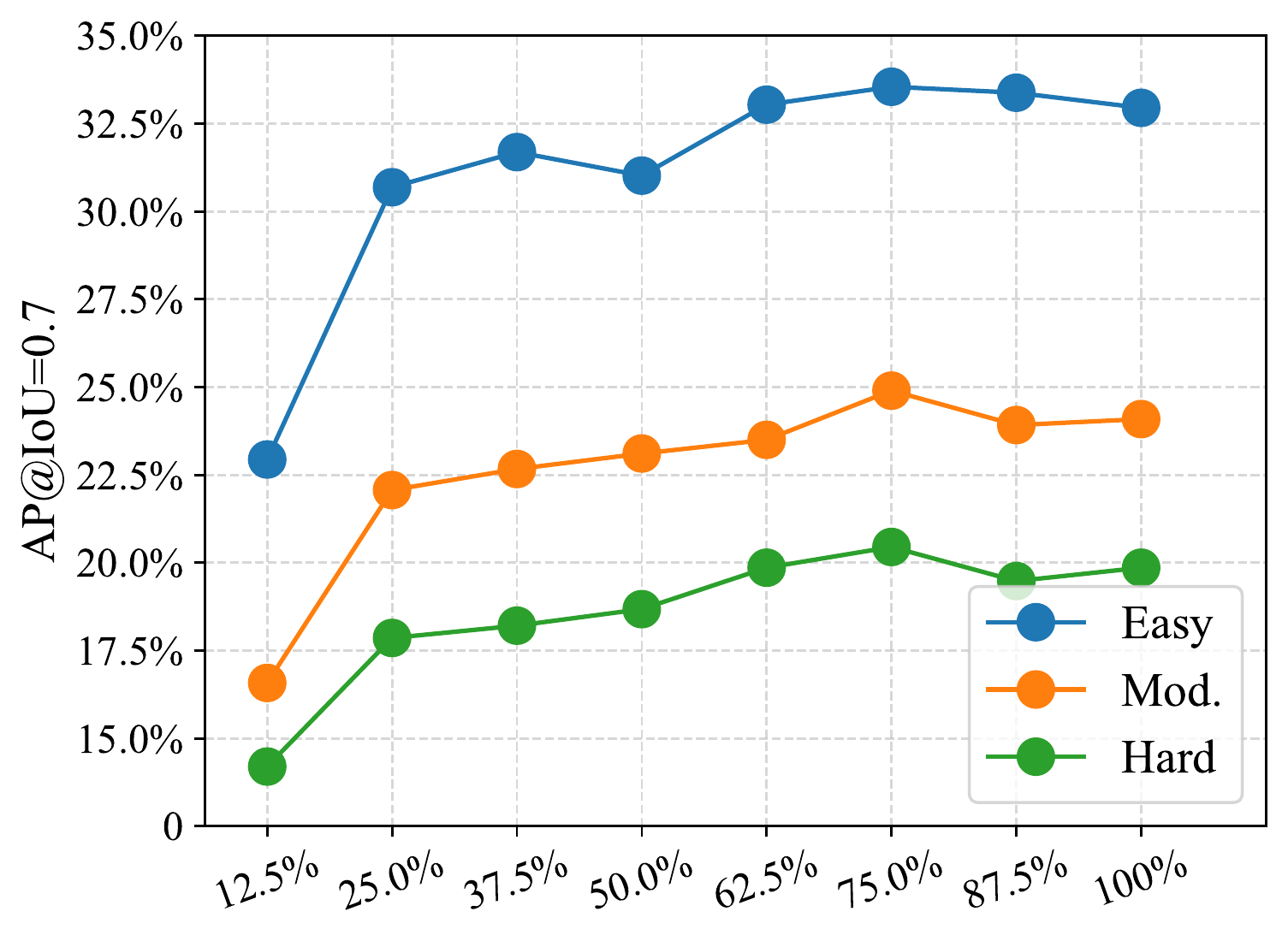}}
{\includegraphics[width=0.4\linewidth]{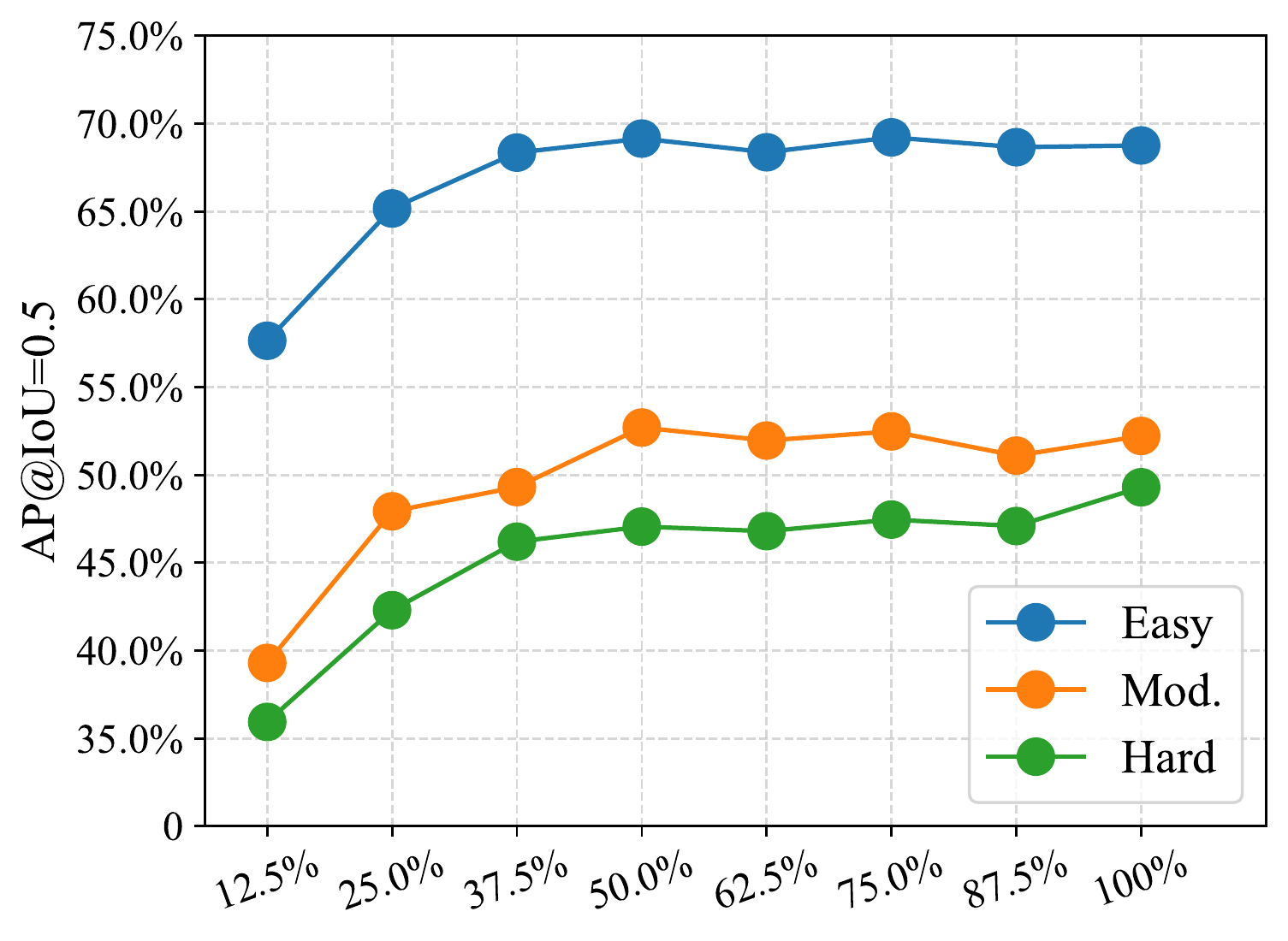}}
\caption{{\bf Sampling rate.} Performance curve of the GUPNet trained
with varying sampling rate. Metrics are the $\rm{AP}|_{R_{40}}$ with 0.5 IoU (\emph{left}) and 0.7 IoU (\emph{right}) thresholds on the KITTI \emph{validation} set.}
\label{fig:varying_data_size}
\end{figure*}

\noindent
{\bf Sampling rate.}
We use all frames in the video sequences as training data in Table \ref{tab:more_samples}. However, a reasonable conjecture is we may not need all the frames because the contents captured by adjacent frames are similar, thus providing limited information. Although previous experiments show that denser sampling rate can make the model more stable, presenting the trade-off between the sampling rate and final performance is still meaningful. For this purpose, we uniformly sample the frames from 28,536 training images (train + eigen-clean) to generate several sub-sets, and evaluate the models trained from these sub-sets on \emph{validation} set. As shown in Figure \ref{fig:varying_data_size}, we can see that about 62.5\% of the data can meet the training needs, and more training data does not bring obviously performance improvement. Note that this conclusion can not be directly applied to other datasets due to the different frame rates and moving speed at the data collecting phase. For reference, the KITTI team collected data at 10 Hz with varying driving speeds.

\begin{figure*}[h]
\centering
{\includegraphics[width=0.4\linewidth]{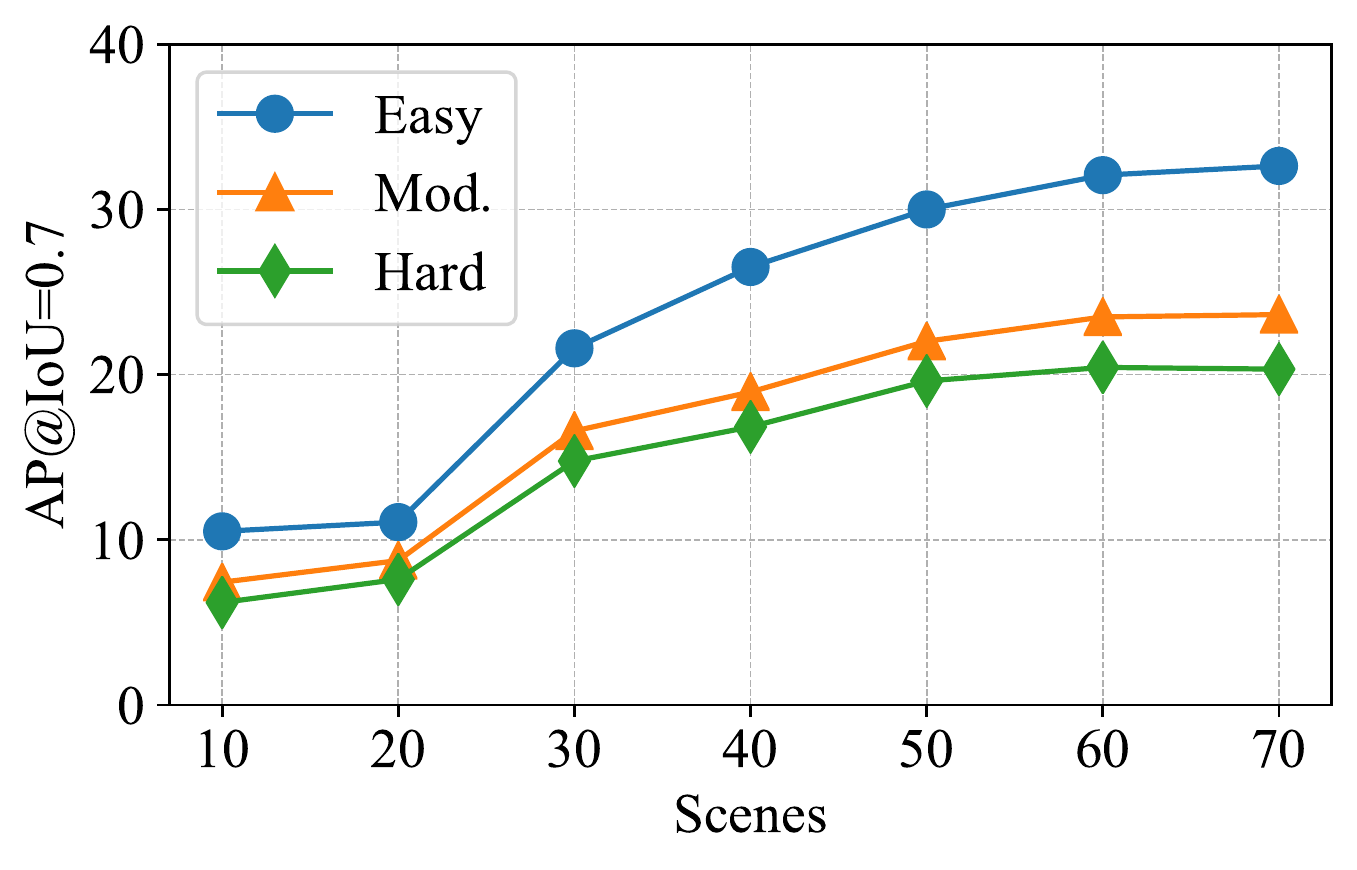}}
{\includegraphics[width=0.4\linewidth]{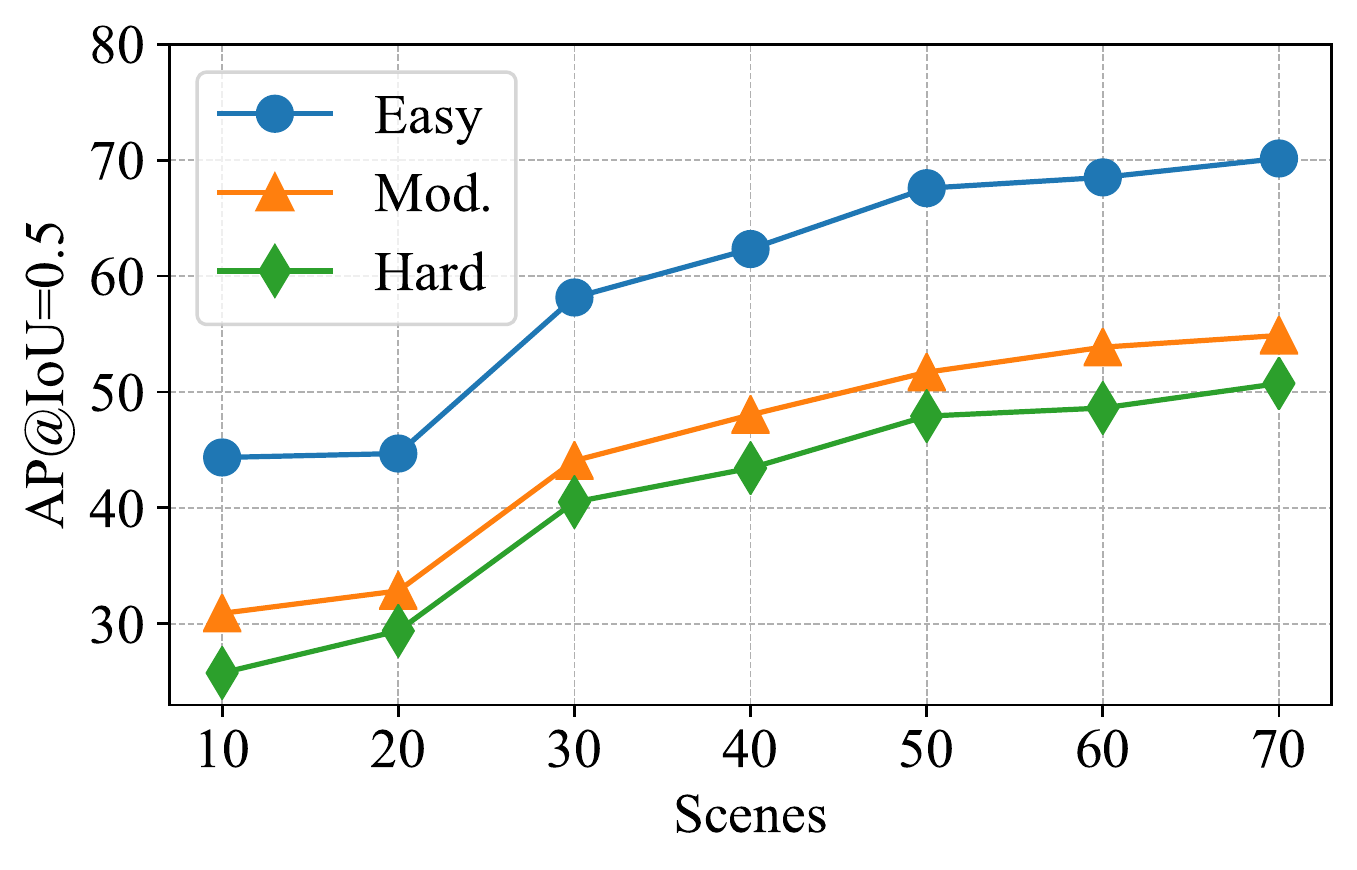}}
\caption{{\bf Scenes diversity.} Performance curve of the GUPNet train with varying scenes diversity. Metrics are the $\rm{AP}|_{R_{40}}$ with 0.5 IoU (\emph{left}) and 0.7 IoU (\emph{right}) thresholds on the KITTI \emph{validation} set.}
\label{fig:scene_diversity}
\end{figure*}

\noindent
{\bf Scenes diversity.}
Compared with denser sampling from the same image set, a more effective way to expand the dataset is collecting more video clips at different scenes. To study the effect of scenes diversity on the performance of monocular 3D detectors, we divide the raw data (train + eigen-clean) into several clips (see Figure \ref{fig:sampling}) and use them to train our models. In particular, each clip contains 200 images (about 20 seconds) and at least 200 images are skipped between adjacent clips.  Figure \ref{fig:scene_diversity} shows the performance changes of the baseline model {\it w.r.t} the increase of the scenes. We can see that the baseline model continues to improve as the increase of the scenes, which indicates the importance of the scenes diversity. We hope these data can provide useful knowledge in data collecting and (pseudo) labeling for future work.

\begin{figure*}[t]
\begin{minipage}[c]{0.55\linewidth}
\centering
\includegraphics[width=0.85\linewidth]{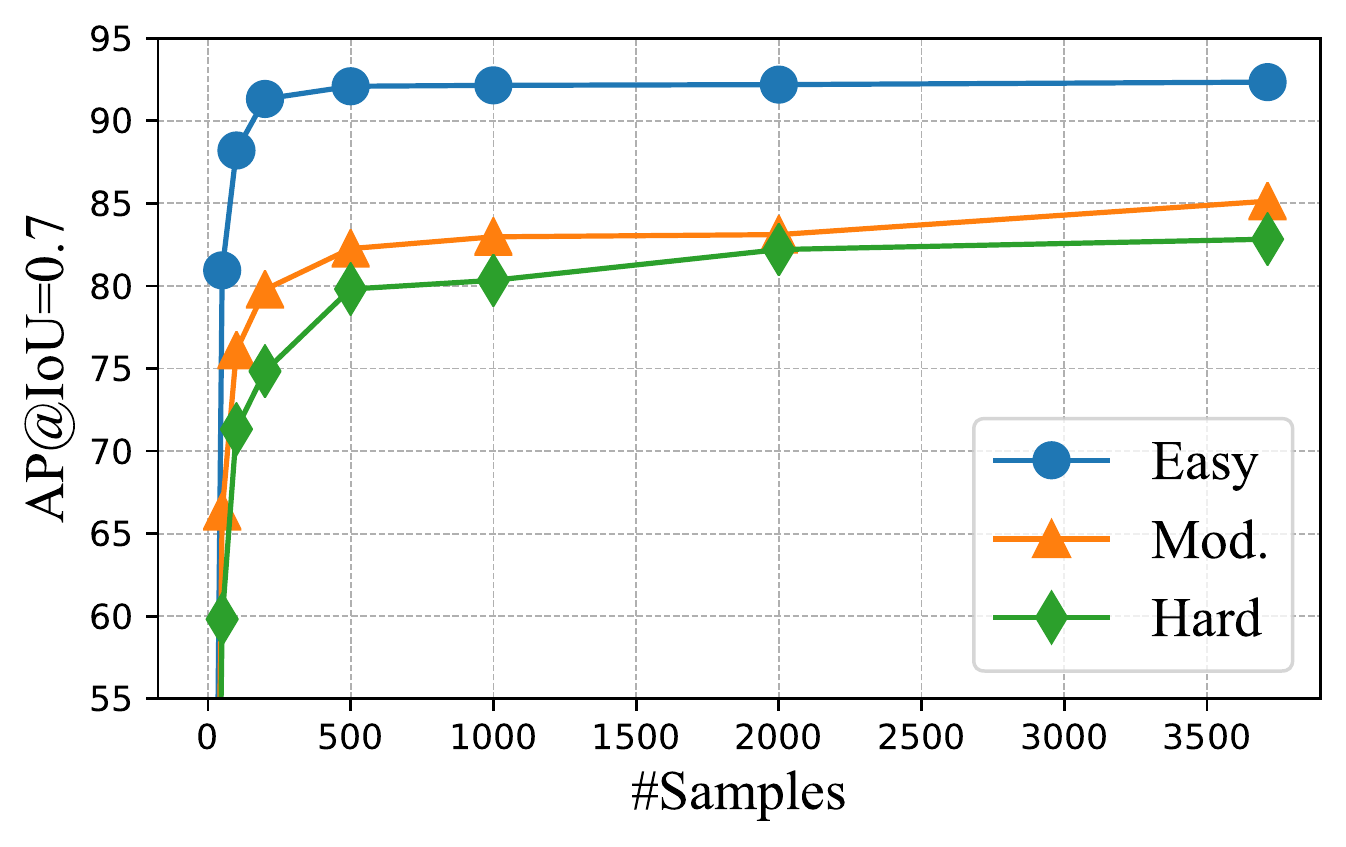}
\end{minipage}
\begin{minipage}[c]{0.45\linewidth}
\small
\setlength\tabcolsep{7.00pt}
\resizebox{0.9\linewidth}{!}{
\begin{tabular}{lr|ccc}
methods & \# imgs & Easy & Mod. & Hard  \\ 
\shline
baseline & 3,712  & 22.37   & 16.21  & 13.98 \\
ours & 50   & 21.13 & 16.25 & 13.66 \\
ours & 100  & 26.86 & 18.84 & 15.96 \\ 
ours & 500  & 26.91 & 19.38 & 16.59 \\
ours & 3,712  & 25.87 & 19.07 & 16.52 \\
\end{tabular}} 
\end{minipage}
\caption{\emph{Left:} the performance curve for Voxel R-CNN with the increase of labeled samples. \emph{Right:} the performance of our method under semi-supervised setting. \# imgs denotes the numbers of \emph{labeled} data, and all models are trained with 3,712 training images.  All models are evaluated on KITTI-3D \emph{validation} set with $\rm{AP}|_{R_{40}}$.}
\label{fig:semi_supervised}
\end{figure*}

\noindent
{\bf Training with less manual annotations.} Except for introducing unlabeled data, another dimension to evaluate the semi-supervised method is reducing the number of labeled samples. Our method also perform good in this setting when the LiDAR signals are available. The results summarized in Figure \ref{fig:semi_supervised} (\emph{left}) demonstrate the effectiveness of our method in semi-supervised setting, where our method obtain comparable performance to the fully supervised baseline/final model with only 50/100 labeled samples. The performance curve shown in Figure \ref{fig:semi_supervised} (\emph{right}) reveals why it works. For the LiDAR-based 3D detectors, the rich and accurate spatial features make the CNN models can learn the mapping function from LiDAR data to 3D results easily, and we can apply lots of augmentations, {\it e.g.} scaling, rotation, shifting, and copy-paste, to effectively extend the training samples for this kind of data. Both of them make the LiDAR-based 3D detectors still work with a few annotations ({\it e.g.} Voxel R-CNN gets 66.35 $\rm{AP}|_{R_{40}}$  with only 50 labeled samples, and reaches `saturation' at about 500 samples). The pseudo-labeling scheme build a bridge between LiDAR-based models and image-based models, and then the latter can also work well under few-shot setting by leveraging LiDAR-based models' feature.

\subsection{More Discussions}
\label{sec:discussion}

\begin{figure*}[t]
\centering
\includegraphics[width=0.48\linewidth]{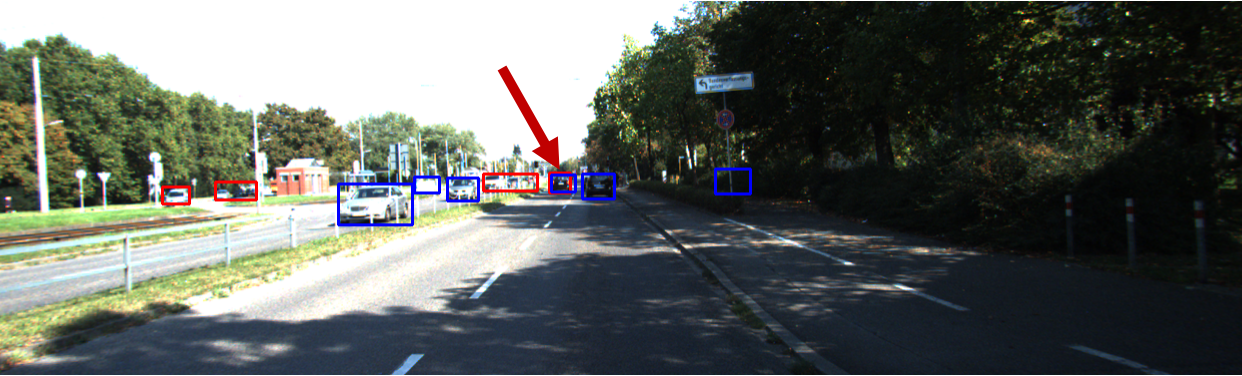}
\includegraphics[width=0.48\linewidth]{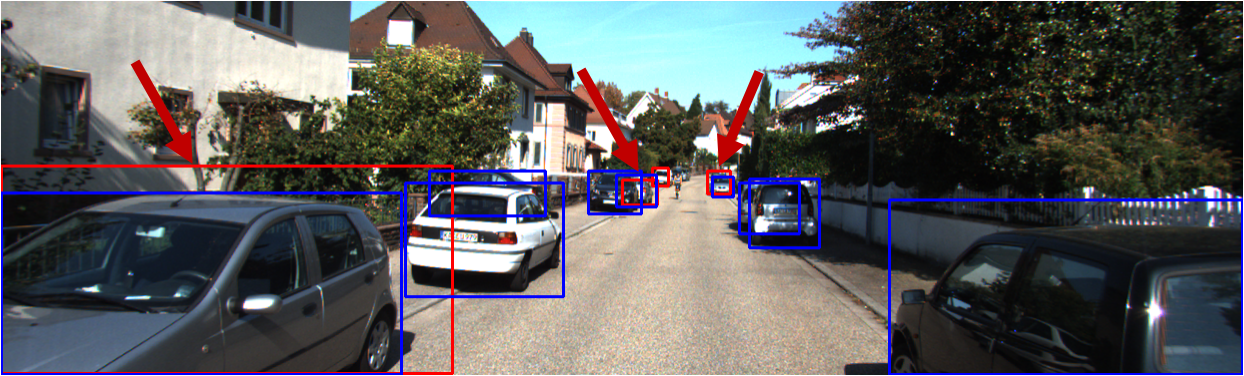}
\includegraphics[width=0.48\linewidth]{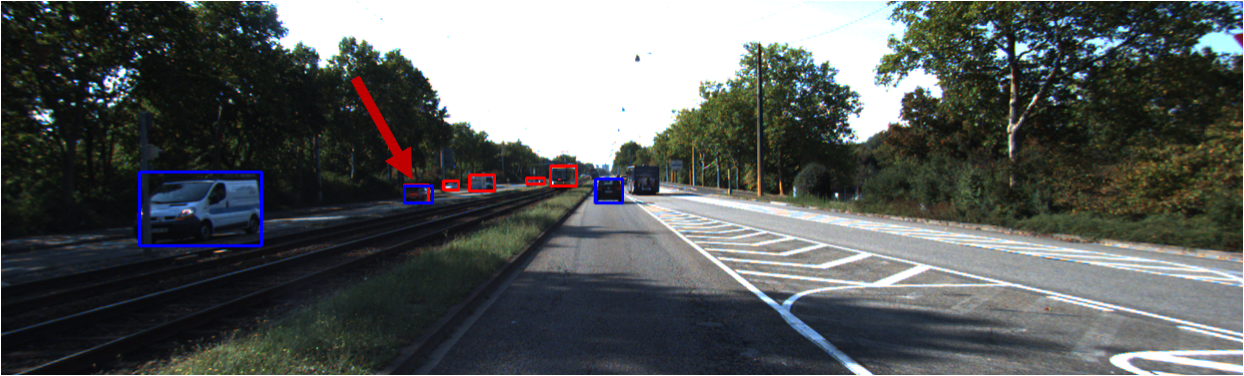}
\includegraphics[width=0.48\linewidth]{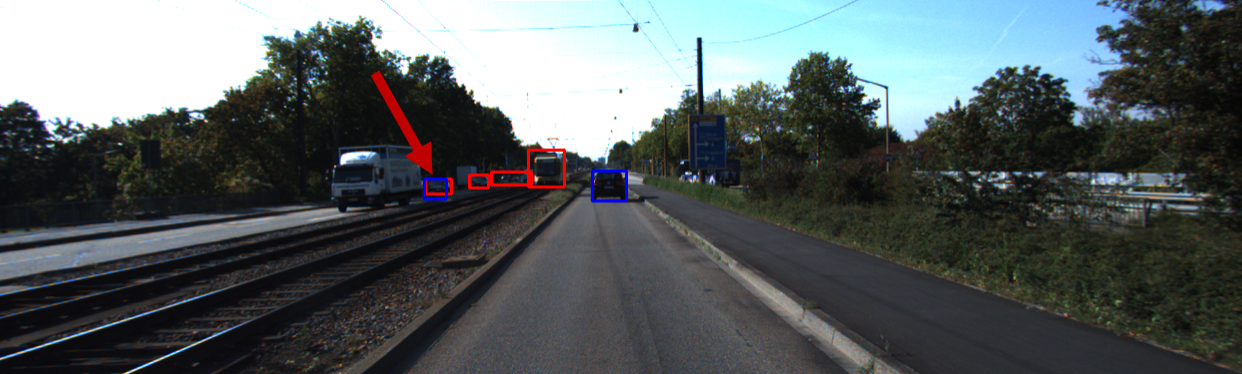}
   \caption{{\bf Comparison of the ground truth and pseudo-labels.} We show the `DontCare' cases in the ground truths in red, and the psuedo-labels in blue. These cases suggest the proposed method can provide more effective supervision than ground truth in some cases. We only show the 2D bounding boxes for clearer presentation.}
   \label{fig:vis}
\end{figure*}

\noindent
{\bf Why pseudo-labeling works?} Here we give an empirical interpretation for the success of pseudo-labeling scheme. First, MonoDLE \cite{monodle} reports that removing some hard (far) instances can boost the models in performance because these samples are hard to detect for monocular detectors and affect CNN’s optimization. The pseudo-labeling also works in a similar way: the teacher model plays the role of the filter to remove the hard instances (MonoDLE uses human designed rules for this purpose), and let the images-based models focus on the remaining samples, which can make the CNN learns the mapping function easier. This suppose is supported by the experiments in Figure \ref{fig:thresholds}: removing the low-confidence labels can improve the model's accuracy. Because the removed samples are not only noised, but also hard to detect. Second, the pseudo-labeling correct some error cases in the annotations (see Figure \ref{fig:vis}). In particular, annotating 3D bounding boxes is a complicated task that involves multi-modal data. For some reason, some objects was not annotated or fully annotated (such as the `DontCare' objects in KITTI-3D, only the 2D bounding boxes are provides, while the 3D 
bounding boxes are missed). For the image data, these objects have the similar texture for the annotated ones, but are regarded as negative samples in the training phase, which misleads the image-based 3D detectors. For our pseudo-labeling scheme, these samples are re-labeled, and then can provide effective supervision to our models. Third, pseudo-labeling does increase the size of training set and leverage the unlabeled data. Lastly, the pseudo-labeling is kind of label-smoothing scheme, which is also beneficial to CNN's optimization.
Based on the above reasons, pseudo-labeling significantly boosts the performances of image-based 3D detectors.

\noindent
{\bf Self-pseudo-labeling.}
In Section \ref{sec:quality} and \ref{sec:scalability}, we show that the pseudo-labels generated from LiDAR/stereo-based methods can provide reliable supervision for monocular 3D detectors (or the LiDAR/stereo combination).
This is reasonable because the pseudo-label generators are generally superior to the baseline methods in performance. 
However, it is questioning whether pseudo-labeling still works without better models?
To study this problem, we generate the pseudo-labels using the GUPNet trained from \emph{training} split, and then train the baseline models using the resulting labels.
Interestingly, as shown in Table \ref{tab:acc_gup}, the models trained on train split and eigen-clean split show contrary results.
Combined with the fact that the \emph{train} split shares the same scenes with the \emph{training} set of pseudo-label generator, we can get the following conclusion: the self-pseudo-labeling scheme still works if the target data share a similar distribution with the labeled data.
In other words, it is still a feasible scheme that manually annotate the key-frames of the raw data and then extend the training set by pseudo-labeling the interframe sequences.
.

\begin{table*}[h]
\centering
\setlength\tabcolsep{5.00pt}
\begin{tabular}{lc|ccc|ccc}
split & \# images & Easy & Mod. & Hard & Easy & Mod. & Hard \\ 
\shline
baseline & 3,712  & 22.37   & 16.21  & 13.98 & 60.50 & 44.97 & 39.99 \\
train & 13,596  & 24.49 & 17.36  &15.31 & 61.28& 45.43 & 40.75\\
eigen-clean & 14,940  & 21.57  & 16.24  & 13.92 &59.05 & 44.69& 38.92 \\
\end{tabular} \vspace{2mm}
\caption{{\bf Pseudo-labeling without LiDAR sweeps.} 
Metrics are the $\rm{AP}_{R_{40}}$ with 0.7/0.5 IoU and 0.5 IoU thresholds.
The baseline model also serve as the pseudo-label generator.}
\label{tab:acc_gup}
\end{table*}

\subsection{Comparing with SOTA Methods}
\label{sec:comparing_sota}

\noindent
{\bf Monocular version.}
We select the results of some representive settings and evaluate them on the KITTI \emph{testing} server.
In particular, the proposed method surpasses DLE~\cite{dle} by 0.54\% for 3D detection, and 1.42\% for BEV under (b) setting; surpasses MonoDistill~\cite{monodistill} by 1.05\% for 3D detection, and 2.19\% for BEV under (c) setting; and surpasses DFR-Net~\cite{dfrnet} 6.6\% for 3D detection, and 8.53\% for BEV, respectively.
These new SOTA performances firmly demonstrate the effectiveness of the proposed methods. See Table \ref{table:mono_sota} for more details.

\noindent
{\bf Stereo version.}
As shown in Table \ref{table:stereo_sota}, to verify the effectiveness of pseudo label, we also submitted our stereo-based results trained based on trainval dataset with pseudo label from Voxel R-CNN~\cite{voxelrcnn} to the official evaluation benchmark for evaluating our performance on the test set of the KITTI dataset.
Under the pure stereo-based setting, we provide \underline{LIGA-Stereo} results by removing the depth loss and imitation loss in LIGA-Stereo~\cite{ligastereo}.
For BEV performance, we surpass RTS3D by 19.44\% mAP. For 3D detection performance, we surpass RTS3D by 21.19\% mAP. Compared with using both stereo images and LiDAR signals in training phase, ours surpass LIGA-Stereo 2.31\% mAP for 3D detection, and 0.62\% for BEV, respectively.

\begin{table*}[!t]
\centering
\setlength\tabcolsep{7.00pt}
\resizebox{0.75\linewidth}{!}{
\begin{tabular}{c|ll|ccc|ccc}
\hline
\multirow{2}{*}{~} &
\multirow{2}{*}{Method} &
\multirow{2}{*}{Venue} & 
\multicolumn{3}{c|}{3D} & \multicolumn{3}{c}{BEV} \\
 ~ & ~ & ~ & Easy & Mod. & Hard  & Easy & Mod. & Hard \\ 
\hline
a & MonoFlex\cite{monoflex} & CVPR'21
& 19.94  & 13.89 & 12.07 
& - & - & - \\
~& AutoShape\cite{autoshape} & ICCV'21
& 22.47 & 14.17 & 11.36 
& 30.66 & 20.08 & 15.59 \\
~ & GUPNet$^{*}$\cite{gupnet} & ICCV'21
& 20.11 & 14.20 & 11.77  
& 30.29 & 21.19 & 18.20\\
~ & MonoCon \cite{monocon} & ICCV'21
& 22.50 & 16.46 & 13.95  
& 31.12 & 22.10 & 19.00\\
\hline
b & RTM3D\cite{rtm3d} & ECCV'20
& 14.41 & 10.34 & 8.77 
& 19.17 & 14.20 & 11.99 \\  
~ & VisualDet3D\cite{groundaware} & RA-L'21
& 21.65 & 13.25 & 9.91 
& 29.81 & 17.98 & 13.08 \\ 
~ & DLE\cite{dle} & BMVC'21
& 24.23 & 14.33 & 10.30 
& 31.09 & 19.05 & 14.13  \\ 
~ & Ours & -
& 23.93 & 14.87 & 12.45
& 33.17 & 20.47 & 17.31 \\ 
\hline
c & MonoPSR\cite{monopsr} & CVPR'19
& 10.76 & 7.25 & 5.85 
& 18.33 & 12.58 & 9.91 \\  
~ & MonoRUn~\cite{monorun} & CVPR'21
& 19.65 & 12.30 & 10.58 
& 27.94 & 17.34 & 15.24\\ 
~ & CaDDN~\cite{caddn} & CVPR'21
& 19.17 & 13.41 & 11.46 
& 27.94 & 18.91 & 17.19 \\ 
~ & MonoDistill~\cite{monodis} & ICLR'22
& 22.97 & 16.03 & 13.60 
& 31.87 & 22.59 & 19.72 \\  
~ & Ours  & ~
& 24.43 & 17.08 & 15.25 
& 33.38 & 24.78 & 22.00 \\   
\hline
d & Demystifying \cite{are_we} & ICCV'21
& 22.40 & 12.53 & 10.64 
& - & - & -  \\  
~ & DDMP-3D \cite{ddmp3d} & CVPR'21
& 19.71 & 12.78 & 9.80 
& 28.08 & 17.89 & 13.44 \\  
~ & PCT\cite{pct} & NeurIPS'21
& 21.00 & 13.37 & 11.31 
& 29.65 & 19.03 & 15.92 \\  
~ & DFR-Net\cite{dfrnet} & ICCV'21
& 19.40 & 13.63 & 10.35
& 28.17 & 19.17 & 14.84 \\  
~ & Ours  & ~
& 28.29 & 20.23 & 17.55 
& 37.81 & 27.70 & 24.61 \\  
\hline
\end{tabular}}
\caption{{\bf Comparing with SOTA methods for monocular setting on KTTI test set.} 
We show the performances of the proposed method and best-performing counterparts under following settings: 
(a) only the monocular images provided by the KITTI-3D are available in the training phase;
(b) both monocular images and stereo images provided by the KITTI-3D are available in the training phase;
(c) both monocular images and LiDAR signals provided by the KITTI-3D are available in the training phase;
(d) both images and LiDAR signals provided by the KITTI 3D and KITTI raw are available in the training phase;
Method are ranked by the $\rm{AP}|_{R_{40}}$ under moderate setting on \emph{testing} set in each group.
$^{*}$: our baseline model.}
\label{table:mono_sota}
\end{table*}

\begin{table*}[!t]
\centering
\setlength\tabcolsep{7.00pt}
\resizebox{0.8\linewidth}{!}{
\begin{tabular}{c|ll|ccc|ccc}
\hline
\multirow{2}{*}{~} &
\multirow{2}{*}{Method} &
\multirow{2}{*}{Venue} & 
\multicolumn{3}{c|}{3D} & \multicolumn{3}{c}{BEV} \\
 ~ & ~ & ~  & Easy & Mod. & Hard  & Easy & Mod. & Hard \\ 
\hline
a  & Stereo R-CNN \cite{stereorcnn} & CVPR'19
& 47.58	& 30.23	& 23.72
& 61.92 & 41.31 & 33.42 \\
~ & SIDE & WACV'19
& 47.69 & 30.82	& 25.68  
& - & - & - \\
~ & Stereo CenterNet & Neurocomputing'22
& 49.94 & 31.30 & 25.62
& - & - & - \\
~ & RTS3D & AAAI'21
& 58.51 & 37.38    & 31.12 
& 72.17 & 51.79 & 43.19 \\
~ & \underline{LIGA-Stereo} (Ours) & ~
& 77.81 & 58.57 & 52.13
& 86.67 & 71.23 & 64.08 \\ 
\hline
b & YOLOStereo3D & ICRA'21
& 65.68	& 41.25	& 30.42
& - & - & - \\
~ & DSGN & CVPR'20
& 73.50	& 52.18	& 45.14
& 82.90 & 65.05 & 56.60 \\
~ & CDN & NeurIPS'20
& 74.52	& 54.22	& 46.36 
& 83.32 & 66.24 & 57.65 \\
~ & LIGA Stereo \cite{ligastereo} & ICCV'21
& 81.39	& 64.66	& 57.22
& 88.15 & 76.78 & 67.40 \\
~ & LIGA-Stereo (Ours) & ~
& 83.77 & 66.97 & 58.41 
& 90.76 & 77.40 & 70.00 \\ 
\hline
\end{tabular}}
\vspace{2mm}
\caption{{\bf Comparing with SOTA methods for stereo setting on KTTI test set.} 
We show the performances of the proposed method and best-performing counterparts under following settings: 
(a) only the stereo images provided by the KITTI-3D in the training phase;
(b) both stereo images and LiDAR signals provided by the KITTI-3D are available in the training phase.
Method are ranked by the $\rm{AP}|_{R_{40}}$ under moderate setting on \emph{testing} set in each group.}
\label{table:stereo_sota}
\end{table*}

\section{Conclusion}
In this work, we present the pseudo-labeling scheme for image-based 3D detection. In this approach, we leverage the side-products in the data collecting and annotating phases, and use these data to generate the pseudo-labels, and then augment the training set of image-based models. Surprisingly, except for the increased size of training data, the pseudo-labels themselves have significantly positive impact on the monocular/stereo models, and we provide empirical explanation for it.
Besides, we also conduct extensive experiments under varying settings to explore the potential scenarios for our method, and our method gets impressive performances for all of them, \emph{i.e.} our method achieves new SOTA for multiple settings on the KITTI-3D \emph{testing} benchmark.
\clearpage

\newpage
\setcounter{section}{0}
\renewcommand\thesection{\Alph{section}} 
\section{Appendix}
\subsection{Details of the Models}
\label{appendix:baselines}
We provide more details about the models used in this work. In particular, we train the LiDAR-based models, {\it i.e.} PV R-CNN \cite{pvrcnn} and Voxel R-CNN \cite{voxelnet}, using their official codebase: OpenPCDet \cite{openpcdet}. For the monocular baselines, we use the official codes provided by MonoDLE \cite{monodle} and GUPNet \cite{gupnet}. Note that, the authors' of MonoDLE adopt the confidence normalization in their open-source code, which makes open-source model performs better than the original version described in the paper by around 1 $\rm{AP}|_{R_{40}}$. For the stereo baseline, we choose the top-performing LIGA-Stereo \cite{ligastereo}, which requires LiDAR points in the training phase. Note that the models in \cite{dsgn,ligastereo} report a stereo-only baseline model which surpasses the existing stereo methods in performance, thus we build a stereo-only baseline by removing the requirements of LiDAR points in LIGA-Stereo, instead of adopting existing stereo models, to evaluate the proposed method under stereo-only setting. We compare the performances of our stereo baseline, \underline{LIGA-Stereo}, and existing methods in Table \ref{tab:stereo}, where suggests our baseline is superior to existing methods in performance, and there are lots of improvement room for existing stereo-only methods.

\begin{table*}[h]
\centering
\setlength\tabcolsep{7.00pt}
\resizebox{0.85\linewidth}{!}{
\begin{tabular}{l|ccc|ccc|ccc|ccc}
\multirow{2}{*}{~} &
\multicolumn{3}{c|}{$\rm{AP}|_{R_{11}}$@3D} & \multicolumn{3}{c|}{$\rm{AP}|_{R_{11}}$@BEV} &
\multicolumn{3}{c|}{$\rm{AP}|_{R_{40}}$@3D} &
\multicolumn{3}{c}{$\rm{AP}|_{R_{40}}$@BEV} \\
method &  Easy & Mod. & Hard & Easy & Mod. & Hard & Easy & Mod. & Hard & Easy & Mod. & Hard\\ 
\shline
Stereo R-CNN\cite{stereorcnn} & 54.11 & 36.69 & 31.07 & 68.50 & 48.30 & 41.47 & - & - & - & - & - & -\\
Stereo CenterNet\cite{stereo_centernet} & - & - & - & - & - & - & 55.25 & 41.44 & 35.13 & 71.26 & 53.27 & 45.53  \\
SIDE\cite{side} & - & - & - & - & - & - & 61.22 & 44.46 & 37.15 & 72.75 & 53.71 & 46.16 \\
RTS3D\cite{rts3d} & - & - & - & - & - & - & 64.76 & 46.70 & 39.27 & 77.50 & 58.65 & 50.14 \\
\underline{LIGA-Stereo} & 76.18  & 57.74  & 54.14 & 86.74 & 68.36 & 65.15 & 77.50 & 58.83  & 52.06 & 88.33 & 70.79 & 63.66\\
\end{tabular}}
\caption{\small{Comparison of the stereo-based 3D detection models on the KITTI \emph{validation} set. We show both $\rm{AP}|_{R_{11}}$ and $\rm{AP}|_{R_{40}}$ for a fair comparison.}}
\label{tab:stereo}
\end{table*}

\subsection{Simulated Sparse LiDAR Signals}
\label{appendix:sparse_lidar}
As we mentioned in Section \ref{sec:quality}, we also evaluate our method with sparse LiDAR points. In particular, following \cite{pseudolidar++}, we generate the simulated 32-beam and 16-beam LiDAR points, and train the LiDAR-based models from the resulting data. We use GUPNet and Voxel R-CNN as the baseline model and PL generator for this part. The experimental results shows that although the resolution of LiDAR signals has a certain impact on the quality of pseudo-labels (see Table \ref{tab:pl_generator}), we can still generate good enough pseudo-labels from the low-resolution LiDAR signals ({\it e.g.} 16-beam)  to train monocular models (see Table \ref{tab:model_under_sparse_lidar}).

\begin{table}[h]
\centering
\setlength\tabcolsep{7.00pt}
\resizebox{\linewidth}{!}{
\begin{tabular}{l|ccc|ccc}
\multirow{2}{*}{~} &
\multicolumn{3}{c|}{$\rm{AP}|_{R_{40}}$@IoU=0.7} & \multicolumn{3}{c}{$\rm{AP}|_{R_{40}}$@IoU=0.5} \\
settings &  Easy & Mod. & Hard &  Easy & Mod. & Hard\\ 
\shline
baseline & 22.37 & 16.21 & 13.98 & 60.50 & 44.97 & 39.99 \\
w/ 64-beam & 25.87 & 19.07 & 16.52 & 64.30 & 48.83 & 44.31 \\
w/ 32-beam & 24.55 & 19.02 & 16.57 & 64.33 & 49.12 & 44.39 \\
w/ 16-beam & 24.62 & 18.44 & 15.25 & 62.32 & 47.08 & 42.37 \\
\end{tabular}}
\caption{\small{Performances of the models trained from the pseudo-labels generated from sparse LiDAR points. Metrics are the $\rm{AP}|_{R_{40}}$ under 0.7 and 0.5 IoU thresholds. All models are trained from 3,712 \emph{training} images.}}
\label{tab:model_under_sparse_lidar}
\end{table}

\subsection{Performances of the PL Generators}
\label{appendix:acc_teachers}
Here we show the performance of the pseudo-label (PL) generators used in this work in Table \ref{tab:pl_generator}. In particular, model (a) is the default PL generator in the experiments part, and model (h) is used to investigate the impact on the student models caused by a different PL generator and show the generalization of the proposed method (Section \ref{sec:quality}). Besides, model (h) is also used to generate the cyclist/pedestrian pseudo-labels (Section \ref{appendix:cyc_ped}). We train models (d, i) to avoid the biased conclusion caused the over-fitting (Section \ref{sec:quality}), and the models (b, c) are used to show our method when the resolution of LiDAR signals is low (Section \ref{appendix:sparse_lidar}). Besides, models (e, f, g) are used to show that our method still performs well when the training samples are limited (Section \ref{sec:scalability}), and models (j, k, l) are serve for the scenarios when LiDAR signals are not available (Section \ref{sec:comparing_sota} and \ref{appendix:stereo}).

\begin{table*}[t]
\centering
\setlength\tabcolsep{7.00pt}
\resizebox{0.65\linewidth}{!}{
\begin{tabular}{c|rlc|ccc}
~ & models & data & setting &  Easy & Mod. & Hard\\ 
\shline
a. & Voxel R-CNN & LiDAR & train $\rightarrow$ val & 92.34 & 85.13 & 82.83 \\
b. & Voxel R-CNN & LiDAR (32-beam) & train $\rightarrow$ val & 92.06 & 80.45& 77.89 \\
c. & Voxel R-CNN & LiDAR (16-beam) & train $\rightarrow$ val &88.51 & 71.96& 69.22  \\
d. & Voxel R-CNN & LiDAR & val $\rightarrow$ train  & 91.85& 82.77 & 77.67 \\
e. & Voxel R-CNN & LiDAR (500 samples) & train $\rightarrow$ val & 92.09 & 82.26 & 79.81 \\
f. & Voxel R-CNN & LiDAR (100 samples) & train $\rightarrow$ val & 88.20 & 76.10 & 71.33 \\
g. & Voxel R-CNN & LiDAR (50 samples) & train $\rightarrow$ val & 80.94 & 66.35 & 59.82  \\
h. & PV R-CNN & LiDAR & train $\rightarrow$ val & 92.17 & 84.53 & 82.38 \\
i. & PV R-CNN & LiDAR & val $\rightarrow$ train  & 92.32 & 82.84 & 77.66 \\
j. & GUPNet & Mono & train $\rightarrow$ val & 23.43 & 17.06 & 14.84 \\
k. & \underline{LIGA Stereo} & Stereo & train $\rightarrow$ val  & 77.05 	& 58.26	 & 51.85 \\
l. & LIGA Stereo & Stereo & train $\rightarrow$ val & 82.32	& 64.29 & 59.34 \\
\end{tabular}}
\caption{Performances of the PL generators used in this work. Metrics are $\rm{AP}|_{R_{40}}$ for 3D detection under 0.7 threshold. Train and val denote the KITTI-3D training and validation set.}
\label{tab:pl_generator}
\end{table*}

\subsection{More Experiments on Stereo Models}
\label{appendix:stereo}

Due to the space limitation, we mainly show the experimental results for the monocular models. Here we provide corresponding experiments to show that the stereo models can also benefit from the larger training set. In particular, just like the monocular models, we train our two stereo baselines on some larger splits with pseudo-labels, and the results are shown in Table \ref{tab:more_samples_stereo_ap40} and \ref{tab:more_samples_stereo_w_lidar_ap40}. 
To provide more reliable and stable results, we report the mean and standard deviation over the last 10 epochs (60 epochs in total). 
All the experiments suggest that more training samples can boost the stereo models in performance. 

\begin{table*}[t]
\centering
\setlength\tabcolsep{5.00pt}
\resizebox{0.9\linewidth}{!}{
\begin{tabular}{lc|lll|lll}
\multirow{2}{*}{~} &
\multirow{2}{*}{~} &
\multicolumn{3}{c|}{$\rm{AP}|_{R_{40}}$@IoU=0.7} & \multicolumn{3}{c}{$\rm{AP}|_{R_{40}}$@IoU=0.5} \\
split & \# images &  Easy & Mod. & Hard &  Easy & Mod. & Hard\\ 
\shline
a. training (gt) & 3,712 & 77.05$_{\pm 0.26}$	& 58.26$_{\pm0.48}$	& 51.85$_{\pm0.13}$	& 97.63$_{\pm1.00}$	    & 89.46$_{\pm0.61}$	& 82.36$_{\pm0.05}$ \\
b. training (pl) & 3,712 & 79.19$_{\pm1.83}$	& 59.61$_{\pm1.23}$	& 54.19$_{\pm1.57}$	& 98.04$_{\pm0.75}$	& 89.62$_{\pm0.86}$	& 84.40$_{\pm0.85}$ \\
c. train & 13,596 & 79.31$_{\pm0.90}$	& 61.20$_{\pm0.88}$	& 55.90$_{\pm1.21}$	& 96.45$_{\pm0.71}$	& 89.91$_{\pm0.17}$	& 84.75$_{\pm0.16}$ \\
d. train $\cup$ eigen-clean & 28,536 & 84.55$_{\pm1.00}$	& 68.72$_{\pm1.26}$	& 63.48$_{\pm0.97}$	& 97.40$_{\pm0.82}$	& 91.96$_{\pm0.22}$	& 88.12$_{\pm0.89}$ \\
\end{tabular}}
\caption{\small{Performances ($\rm{AP}|_{R_{40}}$) on KITTI \emph{validation} set of \underline{LIGA} \underline{Stereo} trained from different data splits. $\pm$ captures the standard deviation over the last 10 epochs. Pseudo-labels are generated by Voxel R-CNN trained from KITTI-3D's \emph{training} split.}}
\label{tab:more_samples_stereo_ap40}
\end{table*}

\begin{table*}[t]
\centering
\setlength\tabcolsep{5.00pt}
\resizebox{0.9\linewidth}{!}{
\begin{tabular}{lc|lll|lll}
\multirow{2}{*}{~} &
\multirow{2}{*}{~} &
\multicolumn{3}{c|}{$\rm{AP}|_{R_{40}}$@IoU=0.7} & \multicolumn{3}{c}{$\rm{AP}|_{R_{40}}$@IoU=0.5} \\
split & \# images &  Easy & Mod. & Hard &  Easy & Mod. & Hard\\ 
\shline
a. training (gt) & 3,712 & 82.32$_{\pm1.15}$	& 64.29$_{\pm1.75}$	& 59.34$_{\pm1.61}$	& 98.76$_{\pm0.78}$	& 92.81$_{\pm0.29}$	& 87.72$_{\pm0.25}$ \\
b. training (pl) & 3,712 & 85.04$_{\pm0.86}$	& 66.77$_{\pm1.03}$	& 61.97$_{\pm0.94}$	& 98.65$_{\pm0.91}$	& 92.78$_{\pm0.29}$	& 89.02$_{\pm0.96}$ \\
c. train & 13,596 & 85.66$_{\pm1.03}$	& 67.22$_{\pm0.51}$	& 62.44$_{\pm0.42}$	& 98.63$_{\pm0.10}$	& 92.70$_{\pm0.13}$	& 89.84$_{\pm0.17}$ \\
d. train $\cup$ eigen-clean & 28,536 & 87.41$_{\pm0.55}$	& 72.98$_{\pm0.56}$	& 68.56$_{\pm0.77}$	& 98.60$_{\pm0.23}$	& 94.63$_{\pm0.15}$	& 91.76$_{\pm0.53}$ \\
\end{tabular}}
\caption{\small{Performances ($\rm{AP}|_{R_{40}}$) on KITTI \emph{validation} set of LIGA Stereo trained from different data splits. $\pm$ captures the standard deviation over the last 10 epochs. Pseudo-labels are generated by Voxel R-CNN trained from KITTI-3D's \emph{training} split.}}
\label{tab:more_samples_stereo_w_lidar_ap40}
\end{table*}

\subsection{Cyclist and Pedestrian}
\label{appendix:cyc_ped}
Due to the high variances in cyclist/pedestrian detection, previous works mainly focus on the car category and attribute this to the limited training samples of cyclist and pedestrian categories. In this work, we introduce more samples and show the performance changes of cyclist/pedestrian detection. In particular, we use the PV R-CNN (Voxel R-CNN is designed for car detection) to generate pseudo-labels and report the mean and standard deviation over the last 10 epochs (140 epochs in total) in Table \ref{tab:pedcyc}. First, overall, we can see that the stereo models trained with pseudo-labels achieve better performances than the baselines (a$\rightarrow$b, e$\rightarrow$f). However, different from the car category, the pseudo-labels are not always better than ground-truth labels, and this may caused by the following two reasons: (i) LiDAR-based models are not good at detecting small or holed objects, which makes quality of pseudo-labels of cyclists/pedestrians worse than those of cars (see Figure \ref{fig:errorcase} for an error case of cyclist/pedestrian detection); (ii) we directly adopt the hyper-parameter ({\it i.e.} confidence threshold) tuned from car detection, and the models may be enhanced by further fine-tuning.
Second, increasing the training samples also improves the accuracy of cyclist/pedestrian detection (b$\rightarrow$c, f$\rightarrow$g). Third, we can find that the proportion of corresponding samples among training set by comparing experiments in Table \ref{tab:pedcyc} and the statistical information shown in Table \ref{tab:statistics}.

\begin{table*}[t]
\centering
\setlength\tabcolsep{5.00pt}
\resizebox{0.8\linewidth}{!}{
\begin{tabular}{lr|rrr|rrr}
\multirow{2}{*}{~} &
\multirow{2}{*}{~} &
\multicolumn{3}{c|}{$\rm{AP}|_{R_{40}}$@IoU=0.5} & \multicolumn{3}{c}{$\rm{AP}|_{R_{40}}$@IoU=0.25} \\
split & \# images &  Easy & Mod. & Hard &  Easy & Mod. & Hard\\ 
\shline
a. training (gt) & 3,712 & 8.75$_{\pm0.33}$ & 7.00$_{\pm0.46}$ & 5.42$_{\pm0.39}$& 33.96$_{\pm0.33}$ & 26.51$_{\pm0.23}$ & 22.15$_{\pm0.19}$ \\
b. training (pl) & 3,712 & 10.59$_{\pm0.22}$ & 8.29$_{\pm0.16}$ & 6.38$_{\pm0.12}$& 30.08$_{\pm0.58}$ & 26.79$_{\pm0.20}$ & 21.98$_{\pm0.73}$ \\
c. train & 13,596 & 20.08$_{\pm0.32}$ & 15.27$_{\pm0.22}$ & 12.12$_{\pm0.15}$& 41.53$_{\pm0.66}$ & 33.15$_{\pm0.18}$ & 27.04$_{\pm0.13}$ \\
d. eigen-clean & 14,940 & 1.21$_{\pm0.19}$ & 0.98$_{\pm0.15}$ & 0.66$_{\pm0.05}$& 9.00$_{\pm0.52}$ & 7.25$_{\pm0.22}$ & 6.15$_{\pm0.19}$ \\
\hline
e. training (gt) & 3,712 & 8.68$_{\pm0.64}$ & 4.48$_{\pm0.18}$ & 4.08$_{\pm0.36}$& 24.36$_{\pm0.51}$ & 13.86$_{\pm0.26}$ & 12.62$_{\pm0.20}$ \\
f. training (pl) & 3,712 & 8.93$_{\pm0.69}$ & 4.31$_{\pm0.39}$ & 3.90$_{\pm0.18}$& 28.05$_{\pm0.74}$ & 15.44$_{\pm0.42}$ & 14.23$_{\pm0.35}$ \\
g. train & 13,596 & 10.24$_{\pm0.32}$ & 5.58$_{\pm0.24}$ & 4.95$_{\pm0.13}$& 30.01$_{\pm0.81}$ & 17.96$_{\pm0.40}$ & 16.33$_{\pm0.34}$ \\
h. eigen-clean & 14,940 & 5.92$_{\pm0.40}$ & 2.84$_{\pm0.34}$ & 2.61$_{\pm0.15}$& 26.87$_{\pm0.88}$ & 14.07$_{\pm0.46}$ & 12.38$_{\pm0.46}$ \\
\end{tabular}}
\caption{\small{Performances of GUPNet on KITTI \emph{validation} set for Pedestrian (\emph{upper group}) and Cyclist (\emph{lower group}). $\pm$ captures the standard deviation over the last 10 epochs. Pseudo-labels are generated by PV R-CNN trained from KITTI-3D's \emph{training} split.}}
\label{tab:pedcyc}
\end{table*}

\begin{table*}[t]
\centering
\setlength\tabcolsep{9.00pt}
\resizebox{0.7\linewidth}{!}{
\begin{tabular}{l|ccccc}
 ~ & training (gt) & training (pl) & train (pl) & eigen-clean (pl) &  \\ 
\shline
\# images & 955 of 3,712 & 1,045 of 3,712 & 5,173 of 13,596 & 1,451 of 14,940  \\
\# instances & 2,207 & 2,209 & 10,189 & 1,767 \\
\hline
\# images & 514 of 3,712 & 566 of 3,712 & 2,019 of 13,596  & 1,205 of 14,940 \\
\# instances & 734 & 826 & 3,048 & 1,270 \\
\end{tabular}}
\caption{\small{The numbers of images and instances of pedestrians (\emph{upper group)} and cyclists (\emph{lower group}). For pseudo-labels, we only consider the instances with confidence over 0.7.}}
\label{tab:statistics}
\end{table*}

\subsection{Qualitative Results}
\label{appendix:visualization}

\noindent
{\bf Videos.} We run our monocular version model and its baseline on three video sequences from KITTI-3D \emph{validation} set, and provide the video presentation in the supplementary materials. Please refer to the video for the qualitative results (our results are more accurate and \emph{stable} than those of the baseline).

\begin{figure*}[t]
\centering
\includegraphics[width=0.7\linewidth]{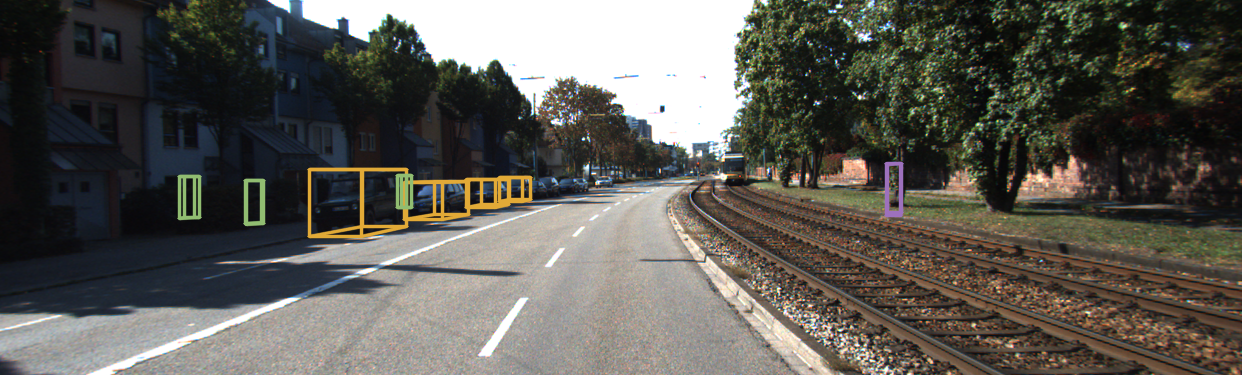}
\includegraphics[width=0.7\linewidth]{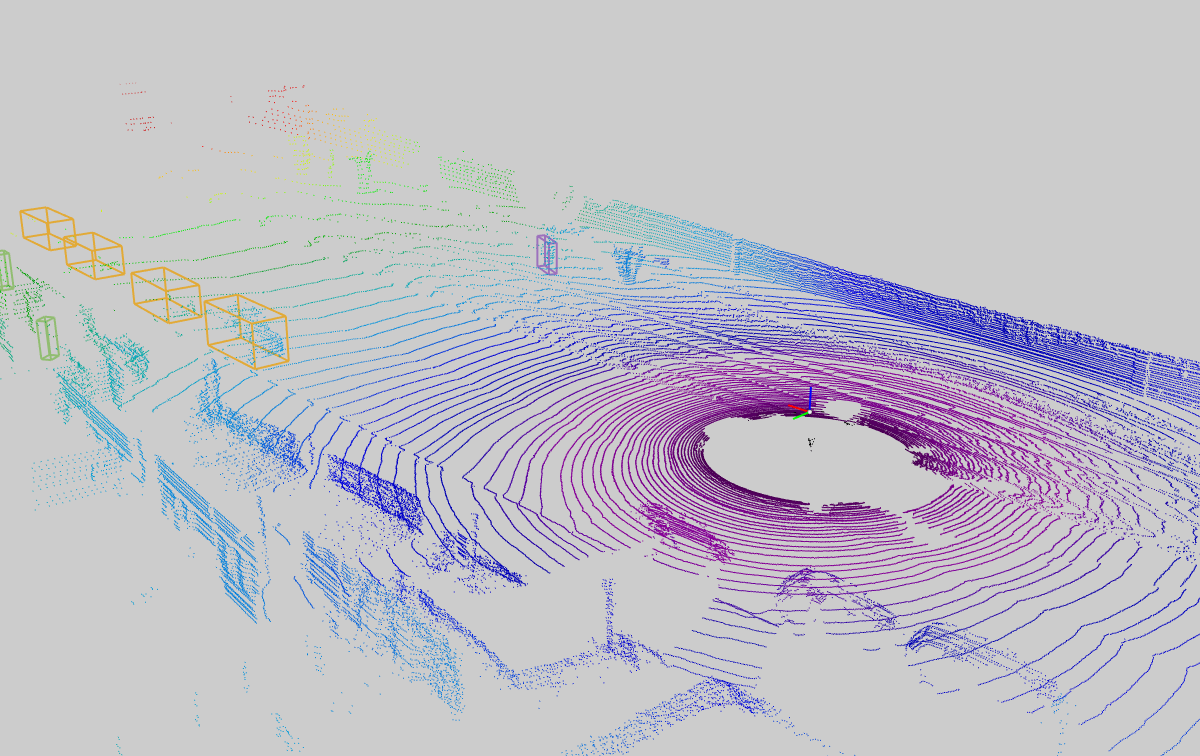}
\caption{An error case in the generated pseudo-labels. We visualize cars, cyclists, and pedestrians with yellow, purple, and green boxes.}
\label{fig:errorcase}
\end{figure*}

\noindent
{\bf Errors case of pseudo-labels.}
In Section \ref{sec:discussion}, we explain why pseudo-labels perform better ground-truth annotations. However, compared with the ground truths, pseudo-labels inevitably introduces some noisy labels, and may mislead the CNNs in training. Here we show a error case in the pseudo-labels generated by PV R-CNN in Figure \ref{fig:errorcase}, where the LiDAR-based models wrongly recognize some background objects as cyclists or pedestrians. These background objects are hard to recognize in 3D space using the point clouds, while are easy to classify with the RGB textures.  

\clearpage

{\small
\bibliographystyle{ieee_fullname}
\bibliography{reference}

\begin{thebibliography}{10}\itemsep=-1pt

\bibitem{m3drpn}
Garrick Brazil and Xiaoming Liu.
\newblock M3d-rpn: Monocular 3d region proposal network for object detection.
\newblock In {\em ICCV}, 2019.

\bibitem{nuscenes}
Holger Caesar, Varun Bankiti, Alex~H Lang, Sourabh Vora, Venice~Erin Liong,
  Qiang Xu, Anush Krishnan, Yu Pan, Giancarlo Baldan, and Oscar Beijbom.
\newblock nuscenes: A multimodal dataset for autonomous driving.
\newblock In {\em CVPR}, 2020.

\bibitem{decoupled3d}
Yingjie Cai, Buyu Li, Zeyu Jiao, Hongsheng Li, Xingyu Zeng, and Xiaogang Wang.
\newblock Monocular 3d object detection with decoupled structured polygon
  estimation and height-guided depth estimation.
\newblock In {\em AAAI}, 2020.

\bibitem{argoverse}
Ming-Fang Chang, John Lambert, Patsorn Sangkloy, Jagjeet Singh, Slawomir Bak,
  Andrew Hartnett, De Wang, Peter Carr, Simon Lucey, Deva Ramanan, and James
  Hays.
\newblock Argoverse: 3d tracking and forecasting with rich maps.
\newblock In {\em CVPR}, 2019.

\bibitem{monorun}
Hansheng Chen, Yuyao Huang, Wei Tian, Zhong Gao, and Lu Xiong.
\newblock Monorun: Monocular 3d object detection by reconstruction and
  uncertainty propagation.
\newblock In {\em CVPR}, 2021.

\bibitem{disprcnn_j}
Linghao Chen, Jiaming Sun, Yiming Xie, Siyu Zhang, Qing Shuai, Qinhong Jiang,
  Guofeng Zhang, Hujun Bao, and Xiaowei Zhou.
\newblock Shape prior guided instance disparity estimation for 3d object
  detection.
\newblock In {\em T-PAMI}, 2021.

\bibitem{mono3d}
Xiaozhi Chen, Kaustav Kundu, Ziyu Zhang, Huimin Ma, Sanja Fidler, and Raquel
  Urtasun.
\newblock Monocular 3d object detection for autonomous driving.
\newblock In {\em CVPR}, 2016.

\bibitem{mv3d}
Xiaozhi Chen, Huimin Ma, Ji Wan, Bo Li, and Tian Xia.
\newblock Multi-view 3d object detection network for autonomous driving.
\newblock In {\em CVPR}, 2017.

\bibitem{dsgn}
Yilun Chen, Shu Liu, Xiaoyong Shen, and Jiaya Jia.
\newblock Dsgn: Deep stereo geometry network for 3d object detection.
\newblock In {\em CVPR}, 2020.

\bibitem{monopair}
Yongjian Chen, Lei Tai, Kai Sun, and Mingyang Li.
\newblock Monopair: Monocular 3d object detection using pairwise spatial
  relationships.
\newblock In {\em CVPR}, 2020.

\bibitem{monodistill}
Zhiyu Chong, Xinzhu Ma, Hong Zhang, Yuxin Yue, Haojie Li, Zhihui Wang, and
  Wanli Ouyang.
\newblock Monodistill: Learning spatial features for monocular 3d object
  detection.
\newblock In {\em ICLR}, 2022.

\bibitem{voxelrcnn}
Jiajun Deng, Shaoshuai Shi, Peiwei Li, Wengang Zhou, Yanyong Zhang, and
  Houqiang Li.
\newblock Voxel r-cnn: Towards high performance voxel-based 3d object
  detection.
\newblock In {\em AAAI}, 2021.

\bibitem{d4lcn}
Mingyu Ding, Yuqi Huo, Hongwei Yi, Zhe Wang, Jianping Shi, Zhiwu Lu, and Ping
  Luo.
\newblock Learning depth-guided convolutions for monocular 3d object detection.
\newblock In {\em CVPR}, 2020.

\bibitem{eigen}
David Eigen, Christian Puhrsch, and Rob Fergus.
\newblock Depth map prediction from a single image using a multi-scale deep
  network.
\newblock In {\em NeurIPS}, 2014.

\bibitem{kitti}
Andreas Geiger, Philip Lenz, and Raquel Urtasun.
\newblock Are we ready for autonomous driving? the kitti vision benchmark
  suite.
\newblock In {\em CVPR}, 2012.

\bibitem{ligastereo}
Xiaoyang Guo, Shaoshuai Shi, Xiaogang Wang, and Hongsheng Li.
\newblock Liga-stereo: Learning lidar geometry aware representations for
  stereo-based 3d detector.
\newblock In {\em ICCV}, 2021.

\bibitem{kd}
Geoffrey Hinton, Oriol Vinyals, Jeff Dean, et~al.
\newblock Distilling the knowledge in a neural network.
\newblock {\em arXiv preprint arXiv:1503.02531}, 2015.

\bibitem{monopsr}
Jason Ku, Alex~D Pon, and Steven~L Waslander.
\newblock Monocular 3d object detection leveraging accurate proposals and shape
  reconstruction.
\newblock In {\em CVPR}, 2019.

\bibitem{groomednms}
Abhinav Kumar, Garrick Brazil, and Xiaoming Liu.
\newblock Groomed-nms: Grouped mathematically differentiable nms for monocular
  3d object detection.
\newblock In {\em CVPR}, 2021.

\bibitem{pointpillars}
Alex~H Lang, Sourabh Vora, Holger Caesar, Lubing Zhou, Jiong Yang, and Oscar
  Beijbom.
\newblock Pointpillars: Fast encoders for object detection from point clouds.
\newblock In {\em Proceedings of the IEEE/CVF Conference on Computer Vision and
  Pattern Recognition}, 2019.

\bibitem{pseudo-label}
Dong-Hyun Lee et~al.
\newblock Pseudo-label: The simple and efficient semi-supervised learning
  method for deep neural networks.
\newblock In {\em ICML Workshop}, 2013.

\bibitem{gs3d}
Buyu Li, Wanli Ouyang, Lu Sheng, Xingyu Zeng, and Xiaogang Wang.
\newblock Gs3d: An efficient 3d object detection framework for autonomous
  driving.
\newblock In {\em CVPR}, 2019.

\bibitem{stereorcnn}
Peiliang Li, Xiaozhi Chen, and Shaojie Shen.
\newblock Stereo r-cnn based 3d object detection for autonomous driving.
\newblock In {\em CVPR}, 2019.

\bibitem{rts3d}
Peixuan Li, Shun Su, and Huaici Zhao.
\newblock Rts3d: Real-time stereo 3d detection from 4d feature-consistency
  embedding space for autonomous driving.
\newblock In {\em AAAI}, 2021.

\bibitem{mono_ssl}
Peixuan Li and Huaici Zhao.
\newblock Monocular 3d detection with geometric constraint embedding and
  semi-supervised training.
\newblock {\em RA-L}, 2021.

\bibitem{rtm3d}
Peixuan Li, Huaici Zhao, Pengfei Liu, and Feidao Cao.
\newblock Rtm3d: Real-time monocular 3d detection from object keypoints for
  autonomous driving.
\newblock In {\em ECCV}, 2020.

\bibitem{dle}
Ce Liu, Shuhang Gu, Luc Van~Gool, and Radu Timofte.
\newblock Deep line encoding for monocular 3d object detection and depth
  prediction.
\newblock In {\em BMVC}, 2021.

\bibitem{monocon}
Xianpeng Liu, Nan Xue, and Tianfu Wu.
\newblock Learning auxiliary monocular contexts helps monocular 3d object
  detection.
\newblock In {\em AAAI}, 2022.

\bibitem{groundaware}
Yuxuan Liu, Yuan Yixuan, and Ming Liu.
\newblock Ground-aware monocular 3d object detection for autonomous driving.
\newblock {\em RA-L}, 2021.

\bibitem{autoshape}
Zongdai Liu, Dingfu Zhou, Feixiang Lu, Jin Fang, and Liangjun Zhang.
\newblock Autoshape: Real-time shape-aware monocular 3d object detection.
\newblock In {\em ICCV}, 2021.

\bibitem{gupnet}
Yan Lu, Xinzhu Ma, Lei Yang, Tianzhu Zhang, Yating Liu, Qi Chu, Junjie Yan, and
  Wanli Ouyang.
\newblock Geometry uncertainty projection network for monocular 3d object
  detection.
\newblock In {\em ICCV}, 2019.

\bibitem{m3dssd}
Shujie Luo, Hang Dai, Ling Shao, and Yong Ding.
\newblock M3dssd: Monocular 3d single stage object detector.
\newblock In {\em CVPR}, 2021.

\bibitem{patchnet}
Xinzhu Ma, Shinan Liu, Zhiyi Xia, Hongwen Zhang, Xingyu Zeng, and Wanli Ouyang.
\newblock Rethinking pseudo-lidar representation.
\newblock In {\em ECCV}, 2020.

\bibitem{3dodi}
Xinzhu Ma, Wanli Ouyang, Andrea Simonelli, and Elisa Ricci.
\newblock 3d object detection from images for autonomous driving: A survey.
\newblock {\em arXiv preprint arXiv:2202.02980}, 2022.

\bibitem{am3d}
Xinzhu Ma, Zhihui Wang, Haojie Li, Pengbo Zhang, Wanli Ouyang, and Xin Fan.
\newblock Accurate monocular 3d object detection via color-embedded 3d
  reconstruction for autonomous driving.
\newblock In {\em ICCV}, 2019.

\bibitem{monodle}
Xinzhu Ma, Yinmin Zhang, Dan Xu, Dongzhan Zhou, Shuai Yi, Haojie Li, and Wanli
  Ouyang.
\newblock Delving into localization errors for monocular 3d object detection.
\newblock In {\em CVPR}, 2021.

\bibitem{deep3dbox}
Arsalan Mousavian, Dragomir Anguelov, John Flynn, and Jana Kosecka.
\newblock 3d bounding box estimation using deep learning and geometry.
\newblock In {\em CVPR}, 2017.

\bibitem{dd3d}
Dennis Park, Rares Ambrus, Vitor Guizilini, Jie Li, and Adrien Gaidon.
\newblock Is pseudo-lidar needed for monocular 3d object detection?
\newblock In {\em ICCV}, 2021.

\bibitem{pl_mono}
Liang Peng, Fei Liu, Zhengxu Yu, Senbo Yan, Dan Deng, Zheng Yang, Haifeng Liu,
  and Deng Cai.
\newblock Lidar point cloud guided monocular 3d object detection.
\newblock {\em arXiv preprint arXiv:2104.09035}, 2021.

\bibitem{side}
Xidong Peng, Xinge Zhu, Tai Wang, and Yuexin Ma.
\newblock Side: Center-based stereo 3d detector with structure-aware instance
  depth estimation.
\newblock In {\em WACV}, 2022.

\bibitem{pseudolidare2e}
Rui Qian, Divyansh Garg, Yan Wang, Yurong You, Serge Belongie, Bharath
  Hariharan, Mark Campbell, Kilian~Q. Weinberger, and Wei-Lun Chao.
\newblock End-to-end pseudo-lidar for image-based 3d object detection.
\newblock In {\em CVPR}, 2020.

\bibitem{monogrnet}
Zengyi Qin, Jinglu Wang, and Yan Lu.
\newblock Monogrnet: A geometric reasoning network for monocular 3d object
  localization.
\newblock In {\em AAAI}, 2019.

\bibitem{tlnet}
Zengyi Qin, Jinglu Wang, and Yan Lu.
\newblock Triangulation learning network: From monocular to stereo 3d object
  detection.
\newblock In {\em CVPR}, 2019.

\bibitem{caddn}
Cody Reading, Ali Harakeh, Julia Chae, and Steven~L Waslander.
\newblock Categorical depth distribution network for monocular 3d object
  detection.
\newblock In {\em CVPR}, 2021.

\bibitem{oftnet}
Thomas Roddick, Alex Kendall, and Roberto Cipolla.
\newblock Orthographic feature transform for monocular 3d object detection.
\newblock In {\em BMVC}, 2019.

\bibitem{pvrcnn}
Shaoshuai Shi, Chaoxu Guo, Li Jiang, Zhe Wang, Jianping Shi, Xiaogang Wang, and
  Hongsheng Li.
\newblock Pv-rcnn: Point-voxel feature set abstraction for 3d object detection.
\newblock In {\em CVPR}, 2020.

\bibitem{pointrcnn}
Shaoshuai Shi, Xiaogang Wang, and Hongsheng Li.
\newblock Pointrcnn: 3d object proposal generation and detection from point
  cloud.
\newblock In {\em CVPR}, 2019.

\bibitem{monorcnn}
Xuepeng Shi, Qi Ye, Xiaozhi Chen, Chuangrong Chen, Zhixiang Chen, and Tae-Kyun
  Kim.
\newblock Geometry-based distance decomposition for monocular 3d object
  detection.
\newblock In {\em ICCV}, 2021.

\bibitem{stereo_centernet}
Yuguang Shi, Yu Guo, Zhenqiang Mi, and Xinjie Li.
\newblock Stereo centernet-based 3d object detection for autonomous driving.
\newblock {\em Neurocomputing}, 2022.

\bibitem{are_we}
Andrea Simonelli, Samuel~Rota Bul\`o, Lorenzo Porzi, Peter Kontschieder, and
  Elisa Ricci.
\newblock Are we missing confidence in pseudo-lidar methods for monocular 3d
  object detection?
\newblock In {\em ICCV}, 2021.

\bibitem{demystifying}
Andrea Simonelli, Samuel~Rota Bul{\`o}, Lorenzo Porzi, Peter Kontschieder, and
  Elisa Ricci.
\newblock Are we missing confidence in pseudo-lidar methods for monocular 3d
  object detection?
\newblock In {\em ICCV}, 2021.

\bibitem{monodis}
Andrea Simonelli, Samuel~Rota Bulo, Lorenzo Porzi, Manuel L{\'o}pez-Antequera,
  and Peter Kontschieder.
\newblock Disentangling monocular 3d object detection.
\newblock In {\em ICCV}, 2019.

\bibitem{waymo}
Pei Sun, Henrik Kretzschmar, Xerxes Dotiwalla, Aurelien Chouard, Vijaysai
  Patnaik, Paul Tsui, James Guo, Yin Zhou, Yuning Chai, Benjamin Caine, et~al.
\newblock Scalability in perception for autonomous driving: Waymo open dataset.
\newblock In {\em CVPR}, 2020.

\bibitem{openpcdet}
OpenPCDet~Development Team.
\newblock Openpcdet: An open-source toolbox for 3d object detection from point
  clouds.
\newblock \url{https://github.com/open-mmlab/OpenPCDet}, 2020.

\bibitem{ioumatch}
He Wang, Yezhen Cong, Or Litany, Yue Gao, and Leonidas~J Guibas.
\newblock 3dioumatch: Leveraging iou prediction for semi-supervised 3d object
  detection.
\newblock In {\em CVPR}, 2021.

\bibitem{ddmp3d}
Li Wang, Liang Du, Xiaoqing Ye, Yanwei Fu, Guodong Guo, Xiangyang Xue, Jianfeng
  Feng, and Li Zhang.
\newblock Depth-conditioned dynamic message propagation for monocular 3d object
  detection.
\newblock In {\em CVPR}, 2021.

\bibitem{pct}
Li Wang, Li Zhang, Yi Zhu, Zhi Zhang, Tong He, Mu Li, and Xiangyang Xue.
\newblock Progressive coordinate transforms for monocular 3d object detection.
\newblock In {\em NeurIPS}, 2021.

\bibitem{pseudolidar}
Yan Wang, Wei-Lun Chao, Divyansh Garg, Bharath Hariharan, Mark Campbell, and
  Kilian~Q Weinberger.
\newblock Pseudo-lidar from visual depth estimation: Bridging the gap in 3d
  object detection for autonomous driving.
\newblock In {\em CVPR}, 2019.

\bibitem{monowithpl}
Xinshuo Weng and Kris Kitani.
\newblock Monocular 3d object detection with pseudo-lidar point cloud.
\newblock In {\em ICCVW}, 2019.

\bibitem{multifusion}
Bin Xu and Zhenzhong Chen.
\newblock Multi-level fusion based 3d object detection from monocular images.
\newblock In {\em CVPR}, 2018.

\bibitem{pseudolidar++}
Yurong You, Yan Wang, Wei-Lun Chao, Divyansh Garg, Geoff Pleiss, Bharath
  Hariharan, Mark Campbell, and Kilian~Q. Weinberger.
\newblock Pseudo-lidar++: Accurate depth for 3d object detection in autonomous
  driving.
\newblock In {\em ICLR}, 2020.

\bibitem{monoflex}
Yunpeng Zhang, Jiwen Lu, and Jie Zhou.
\newblock Objects are different: Flexible monocular 3d object detection.
\newblock In {\em CVPR}, 2021.

\bibitem{monogeo}
Yinmin Zhang, Xinzhu Ma, Shuai Yi, Jun Hou, Zhihui Wang, Wanli Ouyang, and Dan
  Xu.
\newblock Learning geometry-guided depth via projective modeling for monocular
  3d object detection.
\newblock {\em arXiv preprint arXiv:2107.13931}, 2021.

\bibitem{sess}
Na Zhao, Tat-Seng Chua, and Gim~Hee Lee.
\newblock Sess: Self-ensembling semi-supervised 3d object detection.
\newblock In {\em CVPR}, 2020.

\bibitem{sessd}
Wu Zheng, Weiliang Tang, Li Jiang, and Chi-Wing Fu.
\newblock Se-ssd: Self-ensembling single-stage object detector from point
  cloud.
\newblock In {\em CVPR}, 2021.

\bibitem{monoef}
Yunsong Zhou, Yuan He, Hongzi Zhu, Cheng Wang, Hongyang Li, and Qinhong Jiang.
\newblock Monocular 3d object detection: An extrinsic parameter free approach.
\newblock In {\em CVPR}, 2021.

\bibitem{voxelnet}
Yin Zhou and Oncel Tuzel.
\newblock Voxelnet: End-to-end learning for point cloud based 3d object
  detection.
\newblock In {\em CVPR}, 2018.

\bibitem{sgm3d}
Zheyuan Zhou, Liang Du, Xiaoqing Ye, Zhikang Zou, Xiao Tan, Errui Ding, Li
  Zhang, Xiangyang Xue, and Jianfeng Feng.
\newblock Sgm3d: Stereo guided monocular 3d object detection.
\newblock {\em arXiv preprint arXiv:2112.01914}, 2021.

\bibitem{dfrnet}
Zhikang Zou, Xiaoqing Ye, Liang Du, Xianhui Cheng, Xiao Tan, Li Zhang, Jianfeng
  Feng, Xiangyang Xue, and Errui Ding.
\newblock The devil is in the task: Exploiting reciprocal
  appearance-localization features for monocular 3d object detection.
\newblock In {\em ICCV}, 2021.

\end{thebibliography}
}

\end{document}